\documentclass[11pt]{article}
\usepackage[final]{acl}
\usepackage[final]{graphicx}
\usepackage{times}
\usepackage{latexsym}
\usepackage{xspace}

\usepackage{tabularx}
\usepackage{array}
\usepackage{booktabs}
\usepackage{multirow}

\usepackage[T1]{fontenc}
\usepackage[utf8]{inputenc}
\usepackage{microtype}
\usepackage{inconsolata}

\usepackage{amsmath,amssymb}
\usepackage{float}
\usepackage[table]{xcolor}
\usepackage{CJKutf8}
\usepackage[most]{tcolorbox}
\usepackage{enumitem}
\usepackage{hyperref}
\usepackage{needspace}
\newenvironment{zh}{\begin{CJK*}{UTF8}{gbsn}}{\end{CJK*}}
\newcommand{\zhtext}[1]{\begin{zh}#1\end{zh}}

\newtcolorbox{promptbox}[1]{
    enhanced,
    breakable,
    colback=gray!4,
    colframe=black!35,
    colbacktitle=gray!25,
    boxrule=0.4pt,
    arc=1pt,
    left=5pt,
    right=5pt,
    top=5pt,
    bottom=5pt,
    before skip=6pt,
    after skip=8pt,
    fonttitle=\bfseries\small,
    coltitle=black,
    title={#1},
    title after break={#1 (continued)}
}

\setlist[itemize]{leftmargin=*,itemsep=0.5pt,topsep=2pt}
\setlist[enumerate]{leftmargin=*,itemsep=0.5pt,topsep=2pt}

\title{CulMind: Benchmarking Multimodal Understanding and Reasoning in Chinese Cultural Heritage}

\author{
  Zhangwei Cao$^{1}$\quad
  Shuhan Fan$^{2}$\quad
  Yuting Wei$^{1}$\thanks{Corresponding author.}\quad
  Jiajun Zhang$^{3}$\quad \\
  \textbf{
  Yihang Peng$^{4}$\quad
  Qi Meng$^{5}$\quad
  Yangfu Zhu$^{6}$\quad
  Liangbin Yang$^{1}$\quad}\\
  $^{1}$University of International Relations\quad $^{2}$Kedge Business School\\
  $^{3}$Peking University\quad
  $^{4}$Tsinghua University\\
  $^{5}$Beijing University of Posts and Telecommunications\quad
  $^{6}$Capital Normal University\\
}

\newcommand{\benchmark}{\textsc{CulMind}\xspace}
\newcommand{\Reasonbenchmark}{\textsc{CulMind-R}\xspace}
\newcommand{\surfaceTasks}{surface-observation tasks\xspace}
\newcommand{\integrativeTasks}{integrative-reasoning tasks\xspace}

\begin{document}
\maketitle

\begin{abstract}
Evaluating Multimodal Large Language Models (MLLMs) in Chinese Cultural Heritage (CCH) requires fine-grained reasoning over visual, textual, stylistic, and historical clues. 
However, existing CCH benchmarks mainly emphasize final-answer accuracy, while the accuracy and completeness of reasoning processes remain underexplored.
To address this gap, we introduce \textbf{\benchmark{}} and \textbf{\Reasonbenchmark{}}: a high-quality benchmark for multimodal CCH covering 50 tasks from collections of more than 100 museums, and a 24-task reasoning subset that adaptively defines task-specific dimensions for reasoning process evaluation.
To evaluate reasoning quality, we propose \textbf{\textsc{ReaScore}}, a task-adaptive metric that evaluates reasoning by automatically weighting task-relevant dimensions.
Experiments on 14 leading MLLMs reveal a substantial gap between answers and reasoning, especially on challenging tasks. Further analysis shows that task-adaptive dimension selection and weighting better align evaluation results with expert judgments. Overall, our benchmark and metric support a more expert-aligned assessment of CCH understanding and offer a transferable reference for broader evaluations of cultural heritage. We publicly release the data, code, and evaluation scripts at \url{https://github.com/ZevTsao/CulMind} to facilitate reproducible research.
\end{abstract}

\begin{figure*}[!t]
	\centering
	\includegraphics[width=0.98\linewidth]{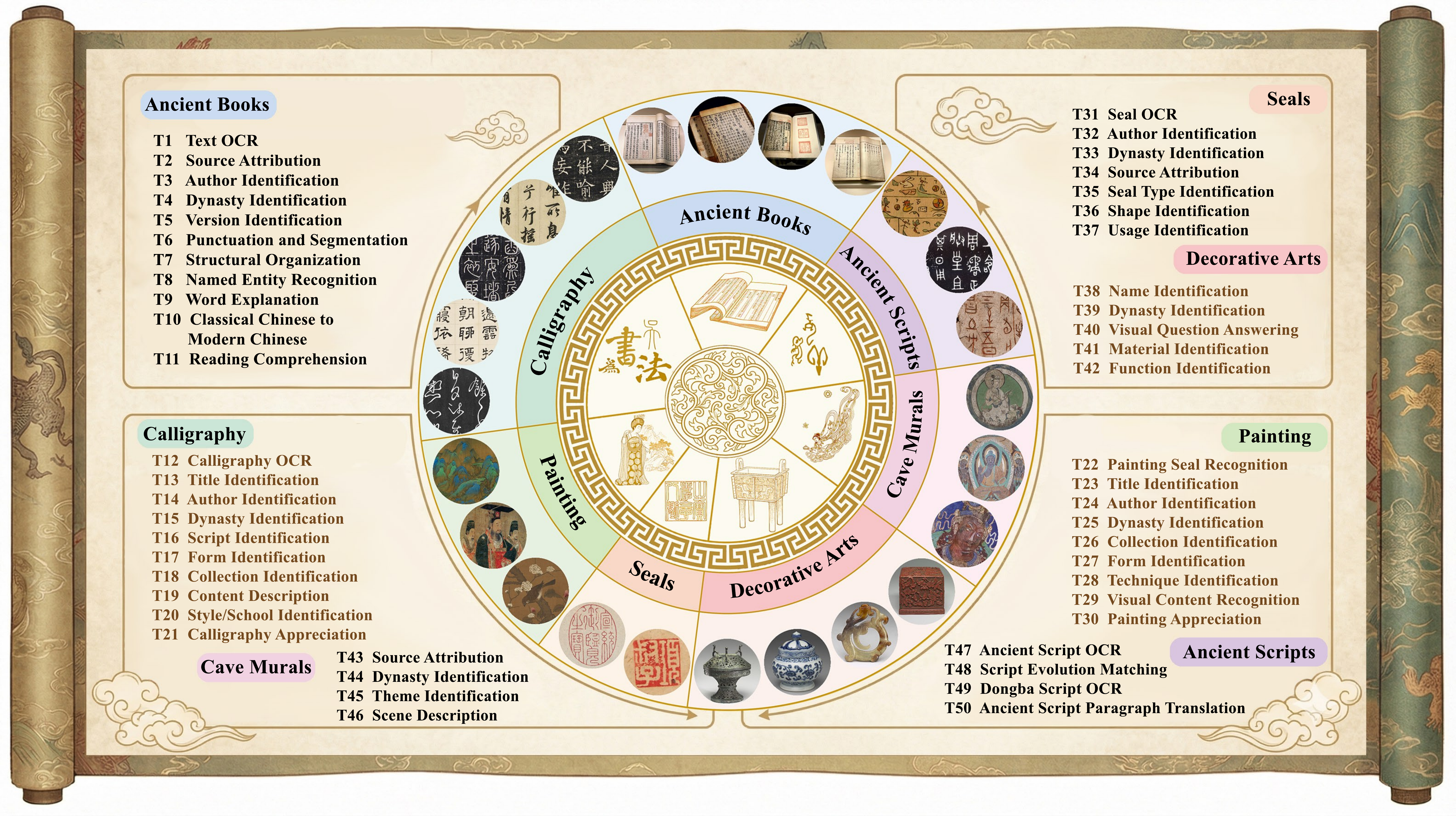}
    \caption{Overview of \benchmark{} and \Reasonbenchmark{}. \benchmark{} covers 50 fine-grained tasks across seven CCH subdomains, while the brown-highlighted region denotes \Reasonbenchmark{}, the 24-task reasoning-process subset. The seven CCH subdomains are illustrated in Figure~\ref{fig:benchmark_subdomains}, and detailed examples of the 50 tasks are provided in Figures~\ref{fig:benchmark_dataset_example_1}--\ref{fig:benchmark_dataset_example_14} in Appendix~\ref{appendix:benchmark_dataset_examples}.}
	\label{fig:pipeline}
\end{figure*}

\section{Introduction}
\label{sec:introduction}
Chinese Cultural Heritage (CCH) preserves millennia of historical memory through diverse multimodal forms, including ancient texts, calligraphy, paintings, seals, decorative arts, cave murals, and ancient characters, making it an important frontier for computational linguistics and digital humanities \citep{zhang2023aclue, zhou2023wyweb, wei2024aceval, cao2024tonggu}. Domain experts interpret CCH by connecting visual details with dynasty, material, form, inscription, seal, brushwork, iconography, provenance, and historical context. Recent advances in Multimodal Large Language Models (MLLMs) have expanded from text understanding to vision-language reasoning and multimodal question answering, creating new opportunities for automated CCH understanding and reasoning \citep{qw25vl, qw3vl, glm45v}. Therefore, assessing the capability of MLLMs to understand and reason over CCH is of significant importance.

General MLLM benchmarks mainly assess broad vision-language capabilities \citep{yue2025mmmupro, fu2025mme, yu2023mmvet, zhang2024cmmmu, he2024cmmu}, while CCH-specific benchmarks extend evaluation to cultural heritage scenarios such as ancient books, paintings, and decorative arts \citep{liu2025mcs, chen2025obibench, yu2025ancientdoc, wei2026cartbench}. Despite their different scopes, these benchmarks largely follow an answer-only-oriented evaluation paradigm, leaving the accuracy and completeness of model reasoning underexamined. 
Process-oriented benchmarks such as REVEAL \citep{jacovi2024reveal}, ProcessBench \citep{zheng2025processbench}, and MME-CoT \citep{jiang2025mmecot} move beyond final answers by verifying intermediate reasoning steps. However, these process-oriented benchmarks still rely on generic reasoning evaluation criteria, making it difficult to capture the task-relevant information dimensions and reasoning requirements in CCH tasks.

To bridge this gap, we propose \textbf{\benchmark{}} and its reasoning subset \textbf{\Reasonbenchmark{}} (Figure~\ref{fig:pipeline}) for evaluating CCH understanding and reasoning processes. \benchmark{} contains 12{,}381 multimodal questions from more than one hundred museums and cultural-heritage institutions, covering 7 subdomains and 50 fine-grained tasks from the Pre-Qin to the Qing dynasty. It balances canonical artefacts with long-tail cultural remains, and includes both surface perception tasks, such as OCR, inscription reading, and visible-object identification, and knowledge-intensive tasks, such as attribution, dating, material inference, stylistic classification, and cultural interpretation. \Reasonbenchmark{} contains 6{,}032 QA pairs and focuses on 24 reasoning-intensive tasks in calligraphy, painting, and decorative arts. Since direct reasoning is difficult to evaluate consistently across heterogeneous CCH tasks, it introduces task-adaptive dimensions to specify the key aspects each task should consider during reasoning. \textbf{\textsc{ReaScore}} then evaluates reasoning quality by adaptively weighting these dimensions and matching them with model-generated reasoning.

Experiments are conducted on 14 mainstream MLLMs, including 10 open-source and 4 closed-source models, under both answer-only and reasoning settings. The results show that 90.6\% of answer-correct instances exhibit correct answers with incorrect reasoning. Human evaluation further shows that \textsc{ReaScore} aligns well with expert assessment, achieving a quadratic weighted kappa (QWK) of 0.9186.

Our contributions are threefold:
\begin{itemize}
    \item We introduce \benchmark{} and its reasoning subset \Reasonbenchmark{} for CCH evaluation. The former covers 50 fine-grained tasks, while the latter focuses on 24 reasoning-intensive tasks with task-adaptive dimensions.

    \item We propose \textsc{ReaScore}, a task-adaptive reasoning evaluation metric that automatically weights task-relevant dimensions.
    
    \item We evaluate 14 MLLMs, including 10 open-source and 4 closed-source models, showing that answer-only evaluation can overlook reasoning errors in CCH understanding.
\end{itemize}

\section{Related Work}
\paragraph{MLLMs for CCH.}
CCH has become an important evaluation setting for MLLMs, as it requires models to jointly recognize visual patterns, textual inscriptions, material attributes, stylistic conventions, and historical context. Existing benchmarks have begun to cover different parts of this space. MCS-Bench \citep{liu2025mcs} evaluates Chinese classical studies across multiple heritage-related subdomains; OBI-Bench \citep{chen2025obibench}, AncientDoc \citep{yu2025ancientdoc}, Museum-65 \citep{balauca2025museum65}, and VQArt-Bench \citep{alfarano2025vqart} focus on oracle bone inscriptions, ancient documents, museum exhibits, and art-oriented cultural-heritage VQA, respectively; while CArtBench \citep{wei2026cartbench} studies Chinese artwork understanding using Palace Museum collections as a single-institution setting.
However, existing benchmarks remain limited in coverage, sources, and attention to long-tail cultural remains. 
Therefore, a systematically designed CCH benchmark is needed to balance domain breadth, source diversity, diachronic span, and long-tail coverage.

\paragraph{Reasoning Benchmarks for MLLMs.}
In general MLLM reasoning evaluation, Learn to Explain \citep{lu2022learntoexplain} introduces thought chains for multimodal science QA; MathVista \citep{lu2024mathvista}, MLLM-CompBench \citep{kil2024mllmcompbench}, NTSEBench \citep{pandya2025ntsebench}, and PhysReason \citep{zhang2025physreason} further evaluate visual mathematical reasoning, comparative reasoning, cognitive reasoning, and physics-based reasoning, respectively. ProJudgeBench \citep{ai2025projudge} evaluates process judging for multimodal scientific reasoning. Compared with general-domain reasoning benchmarks, explicit reasoning evaluation in cultural heritage remains limited. Seeing Culture \citep{satar2025seeingculture} combines cultural visual reasoning with artifact localization; VaseVQA \citep{ge2026vasevqa} shows that domain adaptation can improve cultural-heritage VQA but remains limited on reasoning-intensive questions; and OBI-Bench \citep{chen2025obibench} demonstrates that ancient-script tasks require not only visual perception, but also domain knowledge and rigorous inference.
However, existing benchmarks remain limited in cultural scope, task formats, and reasoning-process design. 
A task-adaptive reasoning benchmark tailored to heterogeneous CCH tasks is still lacking.

\paragraph{Evaluation of Multimodal Reasoning.}
MLLM reasoning evaluation increasingly emphasizes process assessment over final-answer accuracy. Recent general-domain benchmarks such as MME-CoT \citep{jiang2025mmecot}, VisualProcessBench \citep{wang2025visualprm}, and MPBench \citep{xu2025mpbench} evaluate chain-of-thought quality, process supervision, and step-wise correctness in multimodal reasoning.
However, these evaluations often rely on general reasoning criteria and do not account for the task-specific reasoning processes required in cultural-heritage scenarios. Existing classical Chinese benchmarks, such as ACLUE \citep{zhang2023aclue}, WYWEB \citep{zhou2023wyweb}, AC-EVAL \citep{wei2024aceval}, and C$^3$Bench \citep{cao2024c3bench}, are mainly text-oriented and outcome-based. Recent multimodal cultural-heritage benchmarks, including MCS-Bench \citep{liu2025mcs}, AncientDoc \citep{yu2025ancientdoc}, Museum-65 \citep{balauca2025museum65}, VQArt-Bench \citep{alfarano2025vqart}, and CArtBench \citep{wei2026cartbench}, also largely focus on answer accuracy rather than how models identify visual evidence, invoke cultural knowledge, align cross-modal clues, or conduct diachronic inference. This motivates a task-adaptive process-level evaluation mechanism for CCH reasoning.

\section{Data Construction}
\label{sec:data_construction}

\subsection{Task Definition}
\label{sec:task_definition}

\benchmark{} evaluates the multimodal understanding capabilities of MLLMs in CCH, while its reasoning-process subset, \Reasonbenchmark{}, focuses on their process-level reasoning capabilities.
As shown in Figure~\ref{fig:pipeline}, \benchmark{} covers 7 subdomains: \textbf{Ancient Books}, \textbf{Calligraphy}, \textbf{Painting}, \textbf{Seals}, \textbf{Decorative Arts}, \textbf{Cave Murals}, and \textbf{Ancient Scripts}. It comprises 50 expert-designed fine-grained tasks, including OCR, inscription reading, visual content recognition, attribute identification, source attribution, dynasty identification, material and function identification, and cultural interpretation. Detailed subdomain descriptions are provided in Appendix~\ref{appendix:subdomain_details}. The task taxonomy is shown in Figure~\ref{fig:pipeline}.

As the subset, \Reasonbenchmark{} focuses on three selected subdomains: Calligraphy, Painting, and Decorative Arts, and includes 24 reasoning-oriented tasks. We select these 3 subdomains because they represent three major visual-material forms in CCH: calligraphic writing, pictorial images, and decorative artefacts. Together, they cover both directly observable visual cues and integrative judgments involving attribution, dating, material, function, style, and historical context.

\subsection{Data Sources and Item Generation}
\label{sec:data_sources_generation}

In collaboration with scholars in history, philology, art history, and paleography, we select authoritative sources for each CCH subdomain and retain fine-grained metadata from original records. Ancient Books are drawn from manually proofread \textit{Shidian Guji} classics, supplemented with Buddhist and Daoist texts. Visual arts integrate collections from over 100 museums, including the Palace Museum, the British Museum, and the Museum of Fine Arts, Boston. Seals, Cave Murals, and Ancient Scripts draw on specialised resources such as the Fudan seal-impression library, Digital Dunhuang, Chinese Treasure House, paleographic databases, and rare Dongba materials. Full source details are provided in Appendix~\ref{appendix:data_sources}.

We construct \benchmark{} through three item-generation pipelines: expert-written, metadata-instantiated, and LLM-assisted generation. At the corpus level, approximately 30\% of QA items are expert-written, 50\% are instantiated from structured metadata, and 20\% are generated with LLM assistance followed by expert review. Expert-written items target cases requiring visual or contextual interpretation, metadata-instantiated items are derived from authoritative annotations with culturally plausible distractors, and LLM-assisted items are used for description-based and multi-step interpretive questions.

\subsection{Quality Control}
\label{sec:quality_control}

All items are standardized into a unified schema. All generated items are reviewed for answer correctness, image-question consistency, and metadata faithfulness before inclusion. Ambiguous, controversial, or weakly grounded items are revised or removed after expert discussion. Detailed generation, postprocessing, quality-control procedures and qualifications are provided in Appendix~\ref{appendix:generation_quality_control}.

\subsection{\Reasonbenchmark{}: Reasoning-Process Subset}
\label{sec:reasonbench_construction}

For \Reasonbenchmark{}, we further collect and standardize fine-grained metadata for the three selected subdomains to support process-level reasoning evaluation. Since metadata formats differ across institutions and collections, synonymous fields are mapped into a canonical standard format. Detailed field mappings and subdomain-specific field pools are provided in Appendix~\ref{appendix:metadata_standardisation}.

Based on the standardized metadata, we construct an expert-designed reasoning schema for each reasoning-oriented task. Each schema specifies the task-adaptive dimensions a model should report and assigns each dimension one of 4 roles: \textit{target}, \textit{supporting}, \textit{contextual}, or \textit{excluded}. These roles distinguish dimensions that directly answer the task, provide auxiliary information, offer background context, or are not scored.

This role annotation ensures that reasoning-process evaluation is based on task-relevant dimensions rather than a uniform reasoning template. The 24 tasks are further grouped into 6 \surfaceTasks{} and 18 \integrativeTasks{}. Detailed schema construction, role annotation, task grouping, and per-task schema information are provided in Appendix~\ref{appendix:reasonbench_schema}.

\subsection{Data Statistics}
\label{sec:data_statistics}

\benchmark{} contains 12{,}381 QA pairs over 50 tasks, with 13{,}111 underlying raw images. Overall, \benchmark{} includes 9{,}058 multiple-choice items and 3{,}323 open-ended QA items.
\Reasonbenchmark{} contains  6{,}032 QA pairs across 24 reasoning-oriented tasks. These tasks include 6 \surfaceTasks{} and 18 \integrativeTasks{}.

The detailed dataset composition by subdomain, question format, and reasoning-task type is provided in Appendix~\ref{appendix:data_statistics_detailed}.

\section{\textsc{ReaScore}: Task-Adaptive Reasoning-Process Evaluation}
\label{sec:reascore}

\textsc{ReaScore} evaluates structured intermediate predictions against task-relevant reference dimensions. For CCH reasoning, these dimensions include attributes such as inscription, style, material, period, and function, and are scored with task-adaptive weights.

\subsection{Structured Evaluation Object}
\label{sec:reascore_object}

For each instance $x_i$ from task $t$, the model receives an image and a task prompt, and returns a final answer together with a structured set of reasoning-dimension predictions:
\begin{equation}
    y_i = \left(\hat{a}_i, \{\hat{v}_{i,j}\}_{j \in \mathcal{P}_d}\right),
\end{equation}
where $\hat{a}_i$ is the final answer, $\mathcal{P}_d$ is the domain-specific reasoning-dimension pool, and $\hat{v}_{i,j}$ is the prediction for dimension $j$. In \Reasonbenchmark{}, $\mathcal{P}_d$ contains 12 dimensions for Calligraphy, 13 for Painting, and 9 for Decorative Arts. The final answer is scored by the same task-specific answer scorer used in \benchmark{}, while \textsc{ReaScore} focuses on whether the structured intermediate dimensions support that answer. All models are prompted with the same structured output template, and parsing failures or missing fields are handled by deterministic normalization rules in Appendix~\ref{appendix:reascore_field_scoring}.

\subsection{Task-Adaptive Dimension Weighting}
\label{sec:task_adaptive_dimension_weighting}

Each \Reasonbenchmark{} task has an expert-designed schema that assigns every dimension one of four roles: \textit{target}, \textit{supporting}, \textit{contextual}, or \textit{excluded}. Because the same dimension can be decisive in one task and irrelevant in another, \textsc{ReaScore} assigns task-specific weights instead of using a single global weighting rule. Given the domain pool $\mathcal{P}_d$, the task-specific weight vector is
\begin{equation}
\begin{aligned}
\mathbf{w}^{(t)}
&= \{w^{(t)}_j \mid j \in \mathcal{P}_d\}, \\
w^{(t)}_j &\geq 0, \qquad
\sum_{j \in \mathcal{P}_d} w^{(t)}_j = 1, \\
w^{(t)}_j &= 0
\quad \text{if } j \text{ is excluded}.
\end{aligned}
\end{equation}
The active task dimension set is $\mathcal{P}_t=\{j\in\mathcal{P}_d\mid w^{(t)}_j>0\}$. 
Specifically, we use DeepSeek-V4-Pro, a text-only LLM, to infer task-specific weights from the task description, reasoning-dimension pool, and expert role annotations. As it is used only for weight derivation and does not process multimodal inputs, it is not included as a competing model. The derived weights are evaluated against human expert judgments and alternative weighting schemes in Section~\ref{sec:reascore_weighting_rq3}.
Prompting and validation details are provided in Appendix~\ref{appendix:reascore_weight_generation}.

\subsection{Dimension Scoring and Calibration}
\label{sec:reascore_dimension_scoring}

For each active dimension, the prediction is compared with its reference value using a type-aware scoring kernel:
\begin{equation}
    s_{i,j}=\phi_{\tau(j)}(\hat{v}_{i,j},v_{i,j}),
\end{equation}
where $v_{i,j}$ is the reference value, $\tau(j)$ is the dimension type, and $\phi_{\tau(j)}$ dispatches to the appropriate matcher for categorical, textual, list, or numeric values. The full dispatch table and normalization rules are in Appendix~\ref{appendix:reascore_field_scoring}.

CCH images may not provide enough information to determine every dimension. To discourage fabricated specificity, \textsc{ReaScore} calibrates uncertain and missing predictions as follows:
\begin{equation}
\tilde{s}_{i,j}=
\begin{cases}
0.30, & u(\hat{v}_{i,j}) \land v_{i,j}\neq\varnothing,\\
0.50, & u(\hat{v}_{i,j}) \land v_{i,j}=\varnothing,\\
\bot, & \neg u(\hat{v}_{i,j}) \land v_{i,j}=\varnothing,\\
s_{i,j}, & \text{otherwise},
\end{cases}
\end{equation}
where $u(\cdot)$ indicates explicit uncertainty and $\bot$ means that the dimension is removed from the denominator. 
This calibration gives limited credit to honest uncertainty while avoiding credit for unsupported concrete claims. Diagnostics for the 0.30/0.50  calibration constants are reported in Appendix~\ref{appendix:reascore_uncertainty}.

\subsection{Score Aggregation and Consistency Analysis}
\label{sec:reascore_aggregation}

Let $\mathcal{A}_{i,t}=\{j\in\mathcal{P}_t\mid \tilde{s}_{i,j}\neq\bot\}$ denote the active dimensions with valid calibrated scores. The instance-level reasoning-process score is computed as:
\begin{equation}
    R_{i,t}=\frac{\sum_{j\in\mathcal{A}_{i,t}} w^{(t)}_j\tilde{s}_{i,j}}
    {\sum_{j\in\mathcal{A}_{i,t}} w^{(t)}_j}.
\end{equation}

We macro-average over the 24 reasoning tasks to avoid over-weighting tasks with more instances:
\begin{equation}
R =
\frac{1}{|\mathcal{T}_R|}
\sum_{t\in\mathcal{T}_R}
\frac{1}{N_t}
\sum_{i=1}^{N_t} R_{i,t}.
\end{equation}

\section{Experimental Setup}
\label{sec:experimental_setup}

We evaluate 14 mainstream MLLMs. Locally deployed open-source models are served with SGLang in BF16 precision on a single NVIDIA H100 80GB GPU, while the remaining models are accessed through official chat-completion APIs. Full model and deployment details are listed in Appendix~\ref{appendix:model_details}.

All models receive the same image inputs and task-formatted prompts. 
For \benchmark{}, we use an answer-only setting in which models directly produce the final answer for each task.
For \Reasonbenchmark{}, prompts provide the predefined reasoning-dimension pool and require models to return structured dimension-level predictions together with a final answer. 

Each model generates one response per instance without retrieval, web search, or tool use, with temperature set to 0 whenever available. We evaluate \benchmark{} with task-specific answer-only metrics, and evaluate \Reasonbenchmark{} with \textsc{ReaScore} over structured reasoning-dimension predictions. Details on prompts, generation settings, output normalisation, auxiliary LLM-based matching, and result caching are provided in Appendix~\ref{appendix:experimental_details}. Task-specific evaluation metrics are listed in Appendix~\ref{appendix:metrics} and Table~\ref{tab:task_specific_metrics}.
In addition, we use \(\theta_R=0.60\) as the process-reliability threshold on the aggregated instance-level \textsc{ReaScore} to label reasoning processes as reliable or weak. Appendix~\ref{appendix:reascore_threshold_sensitivity} shows that the main conclusions remain stable when this threshold is varied or randomly perturbed.

\section{Results and Analysis}
\label{sec:results_analysis}

We evaluate mainstream MLLMs from two complementary perspectives: answer-only performance on \benchmark{} and reasoning-process reliability on \Reasonbenchmark{}. We organize the analysis around three research questions:
\textbf{RQ1} (\S\ref{sec:main_reasonbenchmark_results}). Do correct final answers reflect reliable reasoning processes in CCH tasks?
\textbf{RQ2} (\S\ref{sec:dimension_selection_rq2}). Are task-adaptive reasoning dimensions necessary for evaluating process reliability?
\textbf{RQ3} (\S\ref{sec:reascore_weighting_rq3}). Is \textsc{ReaScore} a valid measure of reasoning-process reliability?
\begin{table*}[t]
\centering\scriptsize
\setlength{\tabcolsep}{3.2pt}
\renewcommand{\arraystretch}{1.05}
\resizebox{\textwidth}{!}{
\begin{tabular}{lcccccccc}
\toprule
\textbf{Model} & \textbf{Ancient Books} & \textbf{Calligraphy} & \textbf{Painting} & \textbf{Seals} & \textbf{Dec. Arts} & \textbf{Cave Murals} & \textbf{Ancient Scripts} & $\boldsymbol{S}$ \\
\midrule
\rowcolor{gray!15}
\multicolumn{9}{l}{\textbf{Open-source Models}} \\
  Qwen3-VL-8B-Instruct & 0.434 & \textbf{0.651} & \underline{0.480} & 0.483 & 0.812 & 0.362 & \textbf{0.191} & \underline{0.487} \\
  Qwen3-VL-32B-Instruct & \underline{0.458} & 0.607 & 0.423 & \textbf{0.531} & 0.815 & 0.375 & 0.051 & 0.466 \\
  GLM-4.5V & 0.443 & 0.530 & 0.388 & 0.452 & 0.800 & \underline{0.436} & 0.014 & 0.438 \\
  InternVL3-8B & 0.370 & 0.510 & 0.475 & 0.426 & 0.727 & 0.412 & 0.078 & 0.428 \\
  InternVL3-14B & 0.392 & 0.537 & 0.469 & 0.416 & 0.757 & 0.320 & \underline{0.083} & 0.425 \\
  Qwen3.5-27B & 0.396 & 0.500 & 0.390 & 0.410 & 0.730 & 0.312 & 0.040 & 0.397 \\
  Qwen3.5-9B & 0.331 & 0.505 & 0.443 & 0.371 & 0.719 & 0.251 & 0.008 & 0.375 \\
  InternVL3.5-14B-Instruct & 0.333 & 0.488 & 0.415 & 0.272 & 0.594 & 0.341 & 0.068 & 0.359 \\
  InternVL3.5-30B-A3B-Instruct & 0.378 & 0.438 & 0.312 & 0.148 & 0.640 & 0.283 & 0.053 & 0.322 \\
  InternVL3.5-8B-Instruct & 0.362 & 0.441 & 0.298 & 0.238 & 0.552 & 0.235 & 0.045 & 0.310 \\
\midrule
\rowcolor{gray!15}
\multicolumn{9}{l}{\textbf{Closed-source Models}} \\
  Gemini-3-Flash-Preview & \textbf{0.481} & \underline{0.623} & \textbf{0.485} & \underline{0.515} & \textbf{0.892} & \textbf{0.483} & 0.071 & \textbf{0.507} \\
  GPT-5.5 & 0.449 & 0.563 & 0.435 & 0.465 & \underline{0.852} & 0.390 & 0.057 & 0.459 \\
  Gemini-2.5-Flash & 0.434 & 0.564 & 0.452 & 0.454 & 0.820 & 0.363 & 0.044 & 0.447 \\
  GPT-5.4-mini & 0.377 & 0.471 & 0.416 & 0.397 & 0.772 & 0.335 & 0.024 & 0.399 \\
\bottomrule
\end{tabular}
}
\caption{Performance on the \benchmark{} under the answer-only setting. The overall score $S$ is computed by macro-averaging task scores within each subdomain and then averaging over the seven subdomains. \textbf{Bold} denotes the best score, and \underline{underline} denotes the second best.}
\label{tab:culmind_score_results}
\end{table*}

\subsection{Main Results on \benchmark{}}
\label{sec:main_culmind_results}

Table~\ref{tab:culmind_score_results} reports answer-only performance on the full \benchmark{} benchmark. 

\paragraph{Answer-only CCH understanding remains limited and uneven across subdomains.}
As shown in Table~\ref{tab:culmind_score_results}, Gemini-3-Flash-Preview achieves the best overall score of $0.507$, followed by Qwen3-VL-8B-Instruct at $0.487$, the strongest open-source model in our evaluation. However, the leading model varies by subdomain: Gemini-3-Flash-Preview performs best on Ancient Books, Painting, Decorative Arts, and Cave Murals; Qwen3-VL-8B-Instruct leads on Calligraphy and Ancient Scripts; and Qwen3-VL-32B-Instruct achieves the highest score on Seals. This pattern shows that \benchmark{} does not simply measure a single general multimodal ability, but exposes model-specific strengths and weaknesses across different CCH subdomains.

\paragraph{Fine-grained cultural dimensions remain challenging.}
Performance also differs sharply across subdomains. Decorative Arts obtains the highest average score ($0.749$), while Ancient Scripts is the most difficult category ($0.059$). The other subdomains also show clear variation, with average scores ranging from $0.350$ to $0.531$. These results suggest that current MLLMs are more reliable when answers rely on visible categories, shapes, materials, or common decorative patterns, but still struggle with low-frequency scripts, inscriptions, seals, and open-ended cultural interpretation. Model scale alone is also insufficient: Qwen3-VL-8B-Instruct outperforms Qwen3-VL-32B-Instruct in overall $S$, and similar non-monotonic trends appear in the InternVL series. We therefore further evaluate whether models can produce reliable reasoning processes with \Reasonbenchmark{}.

\subsection{Main Results on \Reasonbenchmark{}}
\label{sec:main_reasonbenchmark_results}

\begin{table*}[t]
\centering\scriptsize
\setlength{\tabcolsep}{3.2pt}
\renewcommand{\arraystretch}{1.05}
\resizebox{0.95\textwidth}{!}{
\begin{tabular}{lccccc}
\toprule
\textbf{Model} 
& \textbf{Direct Ans.}$\uparrow$ 
& \textbf{Reasoning Ans.}$\uparrow$ 
& \textbf{\textsc{ReaScore}}$\uparrow$ 
& \begin{tabular}[c]{@{}c@{}}\textbf{Ans. Correct}\\\textbf{Proc. Reliable}$\uparrow$\end{tabular}
& \begin{tabular}[c]{@{}c@{}}\textbf{Ans. Correct}\\\textbf{Proc. Weak}$\downarrow$\end{tabular} \\
\midrule
\rowcolor{gray!15}
\multicolumn{6}{l}{\textbf{Open-source Models}} \\
  Qwen3-VL-32B-Instruct                    & 0.578 & \textbf{0.649} & 0.572 & 0.087 & 0.858 \\
  Qwen3.5-27B                              & 0.504 & 0.596 & 0.585 & 0.062 & 0.892 \\
  GLM-4.5V                                 & 0.530 & 0.583 & 0.552 & 0.057 & 0.876 \\
  Qwen3-VL-8B-Instruct                     & \underline{0.617} & 0.544 & 0.584 & 0.053 & 0.902 \\
  Qwen3.5-9B                               & 0.524 & 0.535 & 0.577 & 0.049 & 0.912 \\
  InternVL3.5-14B-Instruct                 & 0.480 & 0.501 & 0.508 & 0.028 & 0.949 \\
  InternVL3-8B                             & 0.539 & 0.508 & 0.482 & 0.023 & 0.954 \\
  InternVL3-14B                            & 0.555 & 0.513 & 0.480 & 0.022 & 0.963 \\
  InternVL3.5-8B-Instruct                  & 0.408 & 0.414 & 0.501 & 0.018 & 0.963 \\
  InternVL3.5-30B-A3B-Instruct             & 0.430 & 0.458 & 0.457 & 0.013 & 0.978 \\
\midrule
\rowcolor{gray!15}
\multicolumn{6}{l}{\textbf{Closed-source Models}} \\
  Gemini-3-Flash-Preview                   & \textbf{0.624} & 0.613 & \textbf{0.647} & \textbf{0.128} & \textbf{0.776} \\
  GPT-5.5                                  & 0.572 & \underline{0.635} & \underline{0.594} & \underline{0.088} & \underline{0.852} \\
  Gemini-2.5-Flash                         & 0.572 & 0.482 & 0.551 & 0.045 & 0.869 \\
  GPT-5.4-mini                             & 0.509 & 0.546 & 0.533 & 0.028 & 0.940 \\
\bottomrule
\end{tabular}
}
\caption{
Main results on \Reasonbenchmark{}.
\textbf{Direct Ans.} is the answer-only score on the same 24 tasks; \textbf{Reasoning Ans.} is the final-answer score under structured reasoning; and \textbf{\textsc{ReaScore}} measures reasoning-process quality.
\textbf{Ans. Correct / Proc. Reliable} and \textbf{Ans. Correct / Proc. Weak} denote correct-answer cases with reliable and weak reasoning, respectively.
$\uparrow$ / $\downarrow$ indicates higher / lower is better.
\textbf{Bold} and \underline{underline} mark the best and second-best results.
}
\label{tab:reasonbenchmark_main}
\end{table*}

Table~\ref{tab:reasonbenchmark_main} reports answer accuracy and reasoning-process reliability on \Reasonbenchmark{}.

\paragraph{Structured reasoning has mixed effects on final answers.}
Direct Ans. and Reasoning Ans. differ substantially across models, suggesting that structured reasoning prompts do not act as a uniform accuracy booster. Under the structured setting, nine models improve, with the largest gains observed for Qwen3.5-27B (+0.092), Qwen3-VL-32B-Instruct (+0.071), GPT-5.5 (+0.063), and GLM-4.5V (+0.053). Other models degrade: Gemini-2.5-Flash drops by 0.090, and Qwen3-VL-8B-Instruct by 0.073. The best-performing model also changes from Gemini-3-Flash-Preview under direct answering to Qwen3-VL-32B-Instruct under structured reasoning. Thus, asking models to produce intermediate reasoning dimensions can affect final-answer performance, but the direction and magnitude vary by model.

\paragraph{Correct answers often lack reliable reasoning processes.}
\textsc{ReaScore} gives a complementary view beyond final-answer accuracy. Gemini-3-Flash-Preview obtains the highest \textsc{ReaScore} (0.647), followed by GPT-5.5 (0.594). In contrast, Qwen3-VL-32B-Instruct achieves the highest Reasoning Ans. score (0.649), but a lower \textsc{ReaScore} (0.572). This suggests that higher answer accuracy does not necessarily imply better reasoning-process quality.
Furthermore, the instance-level diagnostics show the same pattern. The proportion of Ans. Correct / Proc. Reliable cases is low for all models, with the best value only 0.128 for Gemini-3-Flash-Preview. Meanwhile, Ans. Correct / Proc. Weak remains high, ranging from 0.776 to 0.978. Even for Gemini-3-Flash-Preview, 77.6\% of answer-correct cases are paired with weak reasoning. 
These findings answer \textbf{RQ1} by showing that correct final answers in CCH tasks often fail to reflect reliable reasoning processes.

\subsection{RQ2: Ablation of Task-Adaptive Dimension Selection}
\label{sec:dimension_selection_rq2}

\Reasonbenchmark{} requires models to fill structured reasoning dimensions before producing the final answer. We test the necessity of task-adaptive dimension selection by comparing four sets: \textit{Expert-picked}, an expert-selected reference; \textit{\textsc{ReaScore}}, the default task-adaptive set; \textit{Task-related}, which keeps only direct task dimensions; and \textit{All}, which exposes the full subdomain metadata pool. Dimension-set overlap is computed at the task-schema level over all 24 reasoning tasks. 
For output diagnostics, we sample 20 instances per task, yielding 480 instances per model, and evaluate all 14 MLLMs under each dimension set.
This experiment isolates dimension selection from \textsc{ReaScore} weighting and evaluates each set by expert overlap and output noise.
\begin{table}[t]
\centering
\scriptsize
\setlength{\tabcolsep}{4.5pt}
\renewcommand{\arraystretch}{1.08}
\resizebox{1\columnwidth}{!}{
\begin{tabular}{lccccc}
\toprule
\textbf{Dimension Set} 
& \textbf{Avg. Dims.} 
& \textbf{Prec.} 
& \textbf{Rec.} 
& \textbf{Miss. Rel.} 
& \textbf{Spec. Risk} \\
\midrule
Expert-picked 
& 7.2 & 1.0000 & 1.0000 & 0.0000 & 0.0600 \\
\textsc{ReaScore} 
& 7.3 & 0.9726 & 0.9861 & 0.0139 & 0.0650 \\
Task-related 
& 1.4 & 0.9583 & 0.1969 & 0.8031 & 0.1000 \\
All 
& 11.8 & 0.6183 & 1.0000 & 0.0000 & 0.1800 \\
\bottomrule
\end{tabular}
}
\caption{
Dimension-set comparison.
Avg. Dims. is the average number of selected dimensions.
Expert-picked is used as the reference set for computing precision and recall.
Miss. Rel. denotes the proportion of expert-relevant dimensions not exposed by the configuration.
Spec. Risk measures unsupportedly specific dimension completions.
}
\label{tab:rq2_dimension_selection}
\end{table}

As shown in Table~\ref{tab:rq2_dimension_selection}, the \textsc{ReaScore} dimension set closely approximates the expert-picked reference, achieving 0.9726 precision and 0.9861 recall while using 7.3 dimensions on average. In contrast, the Task-related set is compact but recovers only 19.69\% of expert-relevant dimensions and misses 80.31\% of the required reasoning structure. Direct task dimensions alone are therefore insufficient for CCH reasoning, where inscriptions, seals, visual form, style, material, function, and historical context often provide necessary support. The All set obtains full recall, but its much lower precision indicates that many exposed dimensions are irrelevant to the task.

The output diagnostics show the same trade-off. The Task-related set reduces specificity risk, but largely by removing supporting information. The All set introduces the highest specificity risk, suggesting that exposing the full metadata pool encourages unsupportedly specific completions. The \textsc{ReaScore} dimension set provides a better middle ground: it preserves most expert-relevant reasoning dimensions without requiring models to complete the full metadata pool. Overall, the comparison supports task-adaptive dimension selection as a compact and less noisy interface for process evaluation.

\subsection{RQ3: Validity of \textsc{ReaScore}}
\label{sec:reascore_weighting_rq3}

\textbf{Human evaluation.}
We evaluate whether \textsc{ReaScore} provides a valid measure of reasoning-process reliability. To obtain an external reference, we conduct a blinded expert study on outputs from Gemini-3-Flash-Preview, the best-performing model in our evaluation. We sample 10 instances from each of the 24 reasoning tasks, yielding 240 model-instance outputs, each independently annotated by three CCH experts.

Experts rate each reasoning process along four dimensions: dimension correctness, evidence coverage, inference sufficiency, and uncertainty calibration. The expert process score is the average of the four 1 to 5 Likert scores, and a binary expert-pass label is derived for AUROC and F1 evaluation. Detailed criteria are provided in Appendix~\ref{appendix:expert_annotation}. The annotations show high agreement: ICC(2,3)=0.9656 for continuous process scores, Fleiss' $\kappa=0.8164$ for binary expert-pass labels, and Gwet's AC1=0.8732 as a prevalence-robust agreement measure.

\textbf{Comparison with alternative scoring variants.}
We compare \textsc{ReaScore} with other scoring rules under the same expert reference. All variants use the same frozen structured outputs from Gemini-3-Flash-Preview, isolating the scoring rule rather than generation differences. \textit{Answer Accuracy} uses only final-answer correctness. \textit{Uniform} assigns equal weight to all active reasoning dimensions. \textit{Role-only} uses fixed weights for target, supporting, and contextual roles. \textit{Adaptive}, the full \textsc{ReaScore}, assigns task-specific weights to individual dimensions.

\begin{table}[t]
\centering
\small
\setlength{\tabcolsep}{3.2pt}
\renewcommand{\arraystretch}{1.05}
\resizebox{1\columnwidth}{!}{
\begin{tabular}{lccccc}
\toprule
\textbf{Metric} & $\boldsymbol{\rho}$ & $\boldsymbol{\tau}$ & \textbf{QWK} & \textbf{AUROC} & \textbf{F1} \\
\midrule
Answer Acc. & 0.3277 & 0.2741 & 0.2422 & 0.7718 & 0.6717 \\
Uniform & 0.8596 & 0.6839 & 0.7596 & 0.8550 & 0.7090 \\
Role-only & 0.9723 & 0.8746 & 0.9096 & 0.9861 & 0.9215 \\
Adaptive & \textbf{0.9898} & \textbf{0.9309} & \textbf{0.9186} & \textbf{0.9966} & \textbf{0.9888} \\
\bottomrule
\end{tabular}
}
\caption{
Agreement between automatic metrics and expert process judgments.
$\rho$ and $\tau$ denote Spearman and Kendall correlations; QWK denotes quadratic weighted kappa.
AUROC and F1 are computed against binary expert-pass labels.
}
\label{tab:rq3_reascore_weighting}
\end{table}

As shown in Table~\ref{tab:rq3_reascore_weighting}, Adaptive \textsc{ReaScore} is closest to expert judgments across metrics. Its Spearman correlation reaches 0.9898, compared with 0.9723 for Role-only, 0.8596 for Uniform, and 0.3277 for Answer Accuracy. Uniform performs far better than Answer Accuracy, showing that structured reasoning dimensions capture information missed by final-answer metrics. Role-only further improves over Uniform, suggesting that target, supporting, and contextual dimensions should not be treated equally. Adaptive performs best, indicating that role-level weighting is still too coarse and that dimension value depends on the task.

We use bootstrap resampling to estimate confidence intervals for Spearman correlation improvement. Adaptive improves over Role-only by $\Delta\rho=0.0175$ with a 95\% confidence interval of $[0.0086,0.0315]$, over Uniform by $\Delta\rho=0.1302$ with $[0.0881,0.1907]$, and over Answer Accuracy by $\Delta\rho=0.6621$ with $[0.5007,0.8328]$. These results show that \textsc{ReaScore} is a valid automatic measure of CCH reasoning-process reliability, and that its advantage comes from both explicit reasoning-dimension evaluation and task-adaptive dimension weighting.

\section{Conclusion}
\label{sec:conclusion}

We present \benchmark{}, a multimodal benchmark for CCH, and \Reasonbenchmark{}, a reasoning-process subset. 
We further propose \textsc{ReaScore}, a task-adaptive metric that evaluates structured reasoning processes.
Experiments show that final-answer accuracy can overestimate model reliability, especially on integrative-reasoning tasks. Task-adaptive dimension selection and weighting improve reasoning-process evaluation, and \textsc{ReaScore} aligns better with expert judgments than final-answer accuracy and fixed-weight alternatives.
Overall, our work provides a systematic evaluation pipeline for MLLMs in CCH and offers a reference for assessing multimodal reasoning in other cultural-heritage domains.

\section*{Limitations}
\label{sec:limitations}

Our findings are specific to CCH, whose coverage is constrained by the availability of digitized collections, museum metadata, and expert-verifiable annotations. We mitigate this through multi-source data collection and task-specific evaluation protocols, and the benchmark can be expanded as more high-quality resources become available.
\Reasonbenchmark{} currently focuses on calligraphy, painting, and decorative arts, where structured reasoning dimensions can be defined consistently. Other heritage domains may require different reasoning schemas, but the task-adaptive design allows these dimensions to be revised or expanded without changing the overall evaluation setup.
\textsc{ReaScore} relies on structured dimension matching and process-level assessment, which improves interpretability but may be insufficient for highly open-ended or historically disputed cases. Future extensions can incorporate multiple expert references, uncertainty-aware scoring, and retrieval-augmented CCH reasoning under controlled protocols.

\section*{Ethical Statement}
The benchmark is built from publicly accessible or licensed cultural-heritage resources, including museum catalogues, digitized collections, and scholarly databases, and is intended for non-commercial research and evaluation under the access conditions of the original sources. We reviewed the collected records for personally identifying information and offensive content; names, when present, refer to historical figures, artists, collectors, institutions, or other publicly documented cultural-heritage metadata. CCH annotations are constructed from public records and verifiable references, but some cultural-heritage records may involve uncertainty in attribution, dating, or interpretation. The annotations are therefore intended for model evaluation rather than as final authoritative cultural judgments. The benchmark is intended to evaluate model capabilities, not to support real-world authentication, conservation, provenance assessment, or public interpretation of cultural artifacts without expert review.

\bibliography{custom}

\appendix
\section{Data Construction: Supplementary Details}
\label{appendix:data_construction}

\subsection{Subdomain Descriptions}
\label{appendix:subdomain_details}

\textbf{Ancient Books} refer to books and manuscripts produced before modern printing practices. Digitised images of ancient books preserve not only transcribed or woodblock-printed content, but also paper texture, page layout, binding, typography, and other visual traces that are essential for document-level understanding. Tasks in this subdomain mainly involve OCR, text recognition, title or source identification, and basic content understanding.

\textbf{Calligraphy} is a central Chinese visual art form in which brush, ink, and medium jointly encode textual content, authorship, period style, and cultural context. Our calligraphy images cover handscrolls, hanging scrolls, albums, banners, fan leaves, and rubbings, preserving stroke morphology, ink diffusion, mounting format, paper or silk texture, inscriptions, colophons, and historical damage. Tasks in this subdomain involve inscription reading, script identification, calligraphic style recognition, author or school attribution, and dynasty identification.

\textbf{Painting} refers to traditional Chinese painting on silk, paper, or other carriers, where composition, colour, texture, inscription, and seal evidence jointly support interpretation rather than merely realistic depiction. Digitised paintings therefore retain not only pictorial content but also material texture, colophons, and seals accumulated through circulation history. Tasks in this subdomain involve visual content recognition, subject identification, stylistic interpretation, attribution, dating, and cultural interpretation.

\textbf{Seals} combine calligraphy, spatial composition, and carving technique, serving both as markers of identity or authority and as aesthetic elements in Chinese art. Their digital images preserve seal text, red-white contrast, carving traces, ink-paste texture, side inscriptions, and age-related wear. Tasks in this subdomain involve seal text recognition, inscription reading, style identification, and source or ownership-related interpretation.

\textbf{Decorative Arts} refer to Chinese antiquities such as bronzes, ceramics, porcelain, jade, lacquerware, textiles, and gold or silver objects, whose images capture three-dimensional form, surface ornament, material lustre, and production traces. Tasks in this subdomain involve object recognition, material identification, function inference, technique recognition, dating, and cultural interpretation.

\textbf{Cave Murals} are religious paintings on cave walls and ceilings. High-precision digitisation records scenes, line work, colour, pigment change, flaking, and regional or period-specific artistic features. Tasks in this subdomain involve visual content recognition, iconographic interpretation, motif identification, and cultural or religious interpretation.

\textbf{Ancient Scripts} include early pictographic and ideographic writing systems. In addition to oracle-bone and bronze inscriptions at the origin of Chinese characters, our benchmark also includes Dongba script, a living pictographic writing system used by the Naxi people. Tasks in this subdomain involve character recognition, script-form identification, inscription reading, and interpretation of ancient written forms.

\subsection{Data Sources}
\label{appendix:data_sources}

We collect data from authoritative and publicly accessible CCH resources whenever possible. Source selection is conducted in collaboration with domain scholars in history, philology, art history, and paleography. We prioritise sources with reliable provenance, manually curated metadata, high-quality images, or expert-reviewed textual records.

Ancient Books are mainly drawn from manually proofread \textit{Shidian Guji}\footnote{Shidian Guji: \url{https://www.shidianguji.com/}.} classics, supplemented with Buddhist and Daoist textual resources. For each selected page, we retain available metadata and manually corrected structured text, including edition information, hierarchical titles, paragraph labels, and basic bibliographic attributes. These sources provide reliable textual content and metadata for ancient book understanding tasks.

Visual arts data are integrated from collections of over 100 museums and cultural institutions, including the Palace Museum, the British Museum, and the Museum of Fine Arts, Boston. These sources provide high-resolution images and metadata for calligraphy, painting, decorative arts, and related visual-material categories. The records include titles, authors or makers, dynasties, formats, materials, inscriptions, seals, dimensions, collections, and descriptive notes when available.

Seals, Cave Murals, and Ancient Scripts are drawn from specialised resources such as the Fudan seal-impression library, the Shanghai Library open-data platform, Digital Dunhuang, Chinese Treasure House, paleographic databases, and rare Dongba materials. These resources provide domain-specific materials for seal text recognition, mural understanding, and ancient-script interpretation.

For each source, we retain the original metadata whenever available, including title, creator or maker, dynasty or period, material, technique, object type, dimensions, collection institution, and description. When metadata fields differ across institutions, they are standardised following the procedure described in Appendix~\ref{appendix:metadata_standardisation}.

\subsection{Generation and Quality Control}
\label{appendix:generation_quality_control}
\label{appendix:postprocessing}

We construct \benchmark{} through three item-generation pipelines: expert-written generation, metadata-instantiated generation, and LLM-assisted generation. At the corpus level, approximately 30\% of QA items are expert-written, 50\% are instantiated from structured metadata, and 20\% are produced with LLM assistance followed by expert review.

\textbf{Expert-written items.}
Expert-written items are designed by domain experts for cases that require visual observation, contextual interpretation, or culturally informed judgment. These items cover cases where the answer cannot be obtained by direct metadata conversion, such as description verification, aesthetic appreciation, and culturally grounded interpretation. Researchers design image-grounded questions and verify that each item is representative, answerable, and aligned with real CCH use cases.

\textbf{Metadata-instantiated items.}
Metadata-instantiated items are generated from standardised metadata fields. The correct answer is derived from authoritative source records for fields such as title, author, dynasty, material, object type, collection, format, and seal text. For multiple-choice questions, distractors are sampled from the same field and subdomain, and are further constrained to remain culturally plausible. Distractors may come from similar historical periods, related object types, or visually similar categories.

\textbf{LLM-assisted items.}
LLM-assisted items are used for tasks requiring natural-language context from rich metadata, including description-based and multi-step interpretive questions. Researchers write seed examples and prompts, with a representative template shown in Appendix~\ref{appendix:prompt_templates}; the LLM produces candidate items under strict constraints, and specialists verify factual correctness, answer unambiguity, distractor discriminability, and absence of answer leakage. No LLM-generated item is directly included without expert review. We do not reuse QA pairs from previously released evaluation datasets; all items are constructed from source records, expert-curated metadata, and task-specific annotation protocols.

\textbf{Postprocessing and quality control.}
After generation, all items are converted into a unified data schema with consistent fields for image information, question text, answer format, candidate options when applicable, ground-truth answer, subdomain label, task label, source metadata, and, where applicable, reasoning-process fields. Professional-grade images are compressed to below 1\,MB while preserving details required for fine-grained visual recognition, such as inscription strokes, seal boundaries, brush texture, and decorative patterns. All expert-written and LLM-assisted items are reviewed by domain experts, while metadata-instantiated items undergo deterministic field-consistency checks and stratified expert inspection. Review criteria include answer correctness, image-question consistency, metadata faithfulness, answer uniqueness, absence of answer leakage, and sufficient distinction between correct answers and distractors. Ambiguous, controversial, weakly grounded, or unverifiable items are revised or removed after expert discussion.

\textbf{Annotator and expert qualifications.}
All human annotators and domain experts involved in data construction and verification were recruited from the same laboratory. The team consisted of members with a bachelor's degree or above in Chinese language and literature, art history, or related disciplines, as well as master's-level computer science members with digital humanities research experience. All annotators had more than three years of research experience in the relevant domains and possessed the professional expertise required for the annotation and verification tasks.

\subsection{Prompt Templates}
\label{appendix:prompt_templates}

The original prompts used for LLM-assisted item generation were written in
Chinese. For readability, we present English versions of representative prompt
templates in the paper, while preserving original Chinese cultural terms and
providing English glosses where necessary. Since different subdomains require
different question formats and evidence constraints, we use task-specific prompt
templates rather than a single universal prompt.

\begin{promptbox}{Common Instruction for LLM-assisted Item Generation}
\small

\noindent\textbf{Role.}
You are an expert in Chinese cultural heritage question design, with knowledge of
classical Chinese texts, calligraphy, painting, decorative arts, museum metadata,
and historical interpretation.

\medskip
\noindent\textbf{General Objective.}
Given an image and its associated metadata or textual description, generate a
high-quality evaluation item for the specified task. The question should be
culturally grounded and should require image-based observation, textual
understanding, or domain-specific reasoning, depending on the task type.

\medskip
\noindent\textbf{General Constraints.}
\begin{enumerate}[leftmargin=*, itemsep=1pt, topsep=2pt]
    \item The question stem should be natural, informative, and sufficiently contextualized.
    \item Background information may be provided when it helps situate the object, text, period, genre, or edition.
    \item The stem must not directly reveal the correct answer or contain lexical cues that make the answer obvious.
    \item The answer must be supported by the provided image, source text, metadata, or object description.
    \item Distractors in multiple-choice questions should be culturally plausible and drawn from the same subdomain or task type.
    \item Avoid ambiguous, unverifiable, overly broad, or purely speculative questions.
    \item Do not introduce factual claims that are absent from the input information.
\end{enumerate}
\end{promptbox}

\Needspace{8\baselineskip}
\begin{promptbox}{Ancient Books Prompt Requirements}
\small

For ancient-book tasks, the prompt may provide relevant background such as the
historical period, author, edition, source book, chapter title, and textual
context. For reading-comprehension tasks, the generated item should ask the model
to answer based on the classical Chinese passage and to produce a complete answer
in modern Chinese. The output should contain the question and the reference
answer.
\end{promptbox}

\Needspace{10\baselineskip}
\begin{promptbox}{Calligraphy Prompt Requirements}
\small

For calligraphy tasks, the prompt may provide information about the work, period,
author, format, script type, or cultural background, but it should avoid revealing
the target attribute. Multiple-choice questions contain four options, with no
more than three correct answers. Correct option positions should not follow a
regular pattern.

\begin{itemize}[leftmargin=*, itemsep=1pt, topsep=2pt]
    \item \textbf{Content Description}: identify the textual or semantic content of the work. The stem should avoid directly describing the target content.
    \item \textbf{Style or School Identification}: identify the calligraphic style, school, or tradition, such as Yan style, Liu style, Ou style, Zhao style, running-cursive tradition, stele tradition, or Wu School. The stem should not explicitly name the target style.
    \item \textbf{Calligraphy Appreciation}: identify features of brushwork, structure, rhythm, composition, and artistic mood. The stem may provide historical or textual background, but should not reveal the stylistic answer.
\end{itemize}
\end{promptbox}

\Needspace{10\baselineskip}
\begin{promptbox}{Painting Prompt Requirements}
\small

For painting tasks, the prompt may provide information about period, artist,
format, subject matter, medium, or art-historical context. Multiple-choice
questions contain four options, with no more than three correct answers.

\begin{itemize}[leftmargin=*, itemsep=1pt, topsep=2pt]
    \item \textbf{Technique Identification}: identify techniques such as \zhtext{写意} (\textit{xieyi}, freehand painting), \zhtext{工笔} (\textit{gongbi}, meticulous painting), \zhtext{白描} (\textit{baimiao}, plain line drawing), or blue-and-green coloring. The stem should avoid directly naming the technique.
    \item \textbf{Visual Content Recognition}: identify visible subjects, quantities, or object categories in the painting. The stem should avoid directly revealing the target visual elements.
    \item \textbf{Painting Appreciation}: identify the painting's style, mood, expressive tendency, and aesthetic implications based on the image and metadata.
\end{itemize}
\end{promptbox}

\Needspace{10\baselineskip}
\begin{promptbox}{Decorative Arts Prompt Requirements}
\small

For decorative-arts tasks, the prompt may provide period, collection, material,
shape, glaze, decoration, and appearance, but must not reveal the object's
function. Function-identification items are single-choice questions with four
options. Candidate functions are sampled from the predefined functional label set,
including jewelry, seal, stationery, scholar's object, writing implement,
sculpture, ritual vessel, container, glassware, display object, clothing,
inkstick, flower vessel, food vessel, wine vessel, wooden object, weapon, musical
instrument, incense utensil, and daily-use object. Material labels such as
ceramic, porcelain, jade, lacquer, bronze, gold, and enamel are not treated as
function labels. If the input metadata contains only material labels and no valid
functional label, no function-identification item is generated.
\end{promptbox}

\begin{promptbox}{Example: Ancient Books Reading Comprehension}
\small

\noindent\textbf{Question:}
The \textit{Book of Wei} (\textit{Wei shu}; \zhtext{魏书}) is an important
official history compiled by Wei Shou in the Northern Qi. The selected passage
from ``Basic Annals: Prefatory Record'' (\zhtext{序纪第一}) traces the origin of
the Northern Wei ruling group and describes its early social and cultural
characteristics through a combination of ancient legend and historical memory.
Based on the passage, summarize the early way of life and record-keeping
practices of this group, and explain the cultural background behind the formation
of its surname.

\medskip
\noindent\textbf{Answer:}
The group lived for generations in the vast northern region beyond Youdu. They
were organized under hereditary tribal leaders and relied mainly on animal
husbandry, migration, archery, and hunting. Their customs were simple, and their
mode of social instruction was relatively plain. They had not yet developed a
mature writing system; instead, they recorded affairs through carved marks on
wood and oral transmission, which functioned in a way similar to historical
records. The formation of their surname was associated with the belief that the
Yellow Emperor ruled through the virtue of earth, combined with northern ethnic
linguistic customs that linked the terms for ``earth'' and ``ruler.''
\end{promptbox}

\begin{promptbox}{Example: Calligraphy Appreciation}
\small

\noindent\textbf{Question:}
\textit{Zhongni Mengdian Tie} (\zhtext{仲尼梦奠帖}) is a celebrated
Tang-dynasty calligraphic work. Although modest in scale, it has long been valued
for its refined brushwork, controlled structure, and mature artistic expression.
As a representative work within the running-script tradition, it reflects both
the Tang emphasis on formal discipline and the calligrapher's individual
aesthetic pursuit. Which of the following descriptions best match its artistic
features?

\medskip
\noindent\textbf{Options:}
\begin{itemize}[leftmargin=*, itemsep=1pt, topsep=2pt]
    \item A. The brushwork is vigorous and crisp, with elastic lines and a rigorous yet rhythmic structure.
    \item B. The work relies mainly on dense ink accumulation and strong decorative contrast.
    \item C. The ink tone is restrained and stable, producing a calm and controlled rhythm.
    \item D. The composition is balanced and expansive, showing the natural movement of running script.
\end{itemize}

\noindent\textbf{Answer:} A, C, D
\end{promptbox}

\begin{promptbox}{Example: Painting Technique Identification}
\small

\noindent\textbf{Question:}
In the Ming painting tradition, bird-and-flower subjects were important for both
literati and professional painters. Album leaves were often used to present
small-scale yet refined compositions. \textit{Plum Blossoms and Small Birds}
(\zhtext{梅花小鸟}) depicts natural branches and birds on silk with the use of
color, creating a lively and intimate visual effect. Which painting techniques
are most likely associated with this work?

\medskip
\noindent\textbf{Options:}
\begin{itemize}[leftmargin=*, itemsep=1pt, topsep=2pt]
    \item A. \zhtext{写意} / \textit{xieyi} / freehand painting
    \item B. \zhtext{工笔} / \textit{gongbi} / meticulous painting
    \item C. \zhtext{白描} / \textit{baimiao} / plain line drawing
    \item D. Blue-and-green coloring
\end{itemize}

\noindent\textbf{Answer:} A, B
\end{promptbox}

\begin{promptbox}{Example: Decorative Arts Function Identification}
\small

\noindent\textbf{Question:}
This Yuan-dynasty Longquan celadon object, known as \textit{Longquan Kiln
Celadon Brown-spotted Tripod Flower Container}
(\zhtext{龙泉窑青瓷褐斑三足花囊}), has a basin-like form, a gray body, and a
bluish-green glaze with iron-brown spotted decoration. The exposed areas near the
base and feet show a reddish fired tone, reflecting characteristic Longquan kiln
features. The overall form is stable and distinctive, and the object has appeared
in museum collections and exhibitions. What was its actual function?

\medskip
\noindent\textbf{Options:}
\begin{itemize}[leftmargin=*, itemsep=1pt, topsep=2pt]
    \item A. Flower vessel
    \item B. Container
    \item C. Ritual vessel
    \item D. Display object
\end{itemize}

\noindent\textbf{Answer:} A
\end{promptbox}

\subsection{Metadata Standardisation}
\label{appendix:metadata_standardisation}

For \Reasonbenchmark{}, we collect and standardise fine-grained metadata for Calligraphy, Painting, and Decorative Arts. These metadata fields provide the basis for constructing task-adaptive reasoning schemas.

Metadata formats vary across institutions, collection systems, and subdomains. We therefore map synonymous fields into a canonical standard format. For example, person-related fields such as \textit{creator}, \textit{author}, \textit{artist}, \textit{maker}, and \textit{craftsman} are normalised according to their function in each subdomain.

Temporal fields such as dynasty, period, reign title, and approximate date are also standardised when available. Object-related fields, including object type, format, material, technique, function, dimension, and collection information, are unified across sources.

Uncertain, missing, or ambiguous values are preserved rather than forcibly normalised. This prevents the construction process from introducing unsupported claims into the reasoning evaluation.

The resulting canonical field pools contain 12 fields for Calligraphy, 13 fields for Painting, and 9 fields for Decorative Arts. These fields are used to define task-relevant attributes and task-adaptive reasoning dimensions.

\subsection{\Reasonbenchmark{} Reasoning Schema}
\label{appendix:reasonbench_schema}

For each reasoning-oriented task in \Reasonbenchmark{}, we construct a task-adaptive reasoning schema. The schema defines which dimensions should be considered in a valid reasoning process and how these dimensions relate to the final answer. The initial dimension inventory is derived from the standardised metadata fields of the corresponding subdomain. Domain experts then revise the inventory with reference to scholarly knowledge and task requirements.

Each dimension is assigned one of four roles: \textit{target}, \textit{supporting}, \textit{contextual}, or \textit{excluded}. A \textit{target} dimension directly corresponds to the information required by the question. A \textit{supporting} dimension provides auxiliary information that helps justify the answer. A \textit{contextual} dimension provides useful background information but is not strictly necessary for answering the question. An \textit{excluded} dimension is not scored because it is irrelevant to the task or not expected in a valid reasoning process.

The same field can play different roles across tasks. For example, author is a \textit{target} dimension in author identification, a \textit{supporting} dimension in title identification, and a \textit{contextual} dimension in format identification. In an attribution-related task, the target dimension may be the creator or school, while supporting dimensions may include style, inscription, seal, period, or provenance. In a material-identification task, the target dimension may be the material, while supporting dimensions may include texture, technique, object type, and function.

On average, each task schema contains 7-8 dimensions that are used for scoring: 1-2 target dimensions, 2-4 supporting dimensions, and 2-4 contextual dimensions. Dimensions that are not related to the task are marked as \textit{excluded} and are not taken into account in the scoring process.

Initial independent annotations produced role disagreements in 7.2\% of field--task assignments, mainly at the boundary between supporting and contextual roles. All disagreements are resolved through structured group discussion, and the reconciled schemas are submitted to a senior expert for final audit. The full annotation process spans approximately 2 months.

The 24 reasoning-oriented tasks are grouped into 6 \surfaceTasks{} and 18 \integrativeTasks{}. Surface-observation tasks rely mainly on directly visible image regions, such as inscriptions, seal text, format, or object identity. Integrative-reasoning tasks require combining multiple task-relevant dimensions with domain knowledge, such as attribution, dating, material inference, function inference, stylistic interpretation, and cultural appreciation.

\subsection{Detailed Data Statistics}
\label{appendix:data_statistics_detailed}

\benchmark{} contains 12{,}381 QA pairs over 50 tasks, with 13{,}111 underlying raw images. It includes 9{,}058 multiple-choice items and 3{,}323 open-ended QA items. Table~\ref{tab:data_statistics_detailed} provides the detailed composition of \benchmark{} by subdomain and question format. We report QA-pair counts as the main benchmark scale because a single cultural object may correspond to multiple images, metadata fields, or task-specific references.

\begin{table*}[!htbp]
\centering
\small
\setlength{\tabcolsep}{4.5pt}
\renewcommand{\arraystretch}{1.08}
\begin{tabular}{lrrrrl}
\toprule
\textbf{Subdomain} & \textbf{\#QA} & \textbf{\#MCQ} & \textbf{\#Open-ended} & \textbf{\#Tasks} & \textbf{Time Span} \\
\midrule
Ancient Books & 2{,}748 & 1{,}250 & 1{,}498 & 11 & Pre-Qin--Qing \\
Calligraphy & 2{,}428 & 1{,}998 & 430 & 10 & Han--Qing \\
Painting & 2{,}132 & 1{,}932 & 200 & 9 & Tang--Qing \\
Seals & 1{,}750 & 1{,}500 & 250 & 7 & Warring States--Qing \\
Decorative Arts & 1{,}472 & 1{,}472 & 0 & 5 & Shang--Qing \\
Cave Murals & 851 & 656 & 195 & 4 & Wei--Yuan \\
Ancient Scripts & 1{,}000 & 250 & 750 & 4 & Shang--Qing \\
\midrule
\textbf{Total} & \textbf{12{,}381} & \textbf{9{,}058} & \textbf{3{,}323} & \textbf{50} & \textbf{Shang--Qing} \\
\bottomrule
\end{tabular}
\caption{Dataset composition of \benchmark{} by subdomain. MCQ includes single-choice, multi-choice, and true/false verification questions.}
\label{tab:data_statistics_detailed}
\end{table*}

\Reasonbenchmark{} contains 6{,}032 QA pairs across 24 reasoning-oriented tasks in 3 subdomains: Calligraphy, Painting, and Decorative Arts. The 24 reasoning-oriented tasks include 6 \surfaceTasks{} and 18 \integrativeTasks{}. Surface-observation tasks mainly rely on directly visible image regions, while integrative-reasoning tasks require combining multiple task-relevant dimensions with domain knowledge.

\section{\textsc{ReaScore}: Supplementary Details}
\label{appendix:reascore_details}

\subsection{Type-Aware Process-Field Scoring}
\label{appendix:reascore_field_scoring}

The scoring kernel $\phi_{\tau(j)}$ is selected according to the field type. Table~\ref{tab:reascore_field_scoring_app} summarizes the dispatch rule.

\begin{table}[!htbp]
\centering
\small
\setlength{\tabcolsep}{4pt}
\renewcommand{\arraystretch}{1.08}
\resizebox{1\linewidth}{!}{
\begin{tabular}{lll}
\toprule
\textbf{Field Type} & \textbf{Scoring Kernel} & \textbf{Examples} \\
\midrule
Categorical & match / synonym / semantic / judge & dynasty, author, material \\
Text & BERTScore + ANLS & inscription, colophon \\
List & Set-F1 & seals, attributions \\
Numeric & numeric overlap + ANLS & dimensions \\
\bottomrule
\end{tabular}
}
\caption{Type-aware scoring kernels used by \textsc{ReaScore}.}
\label{tab:reascore_field_scoring_app}
\end{table}

Categorical fields use a cascading matcher: exact match, synonym normalization, substring matching, semantic similarity, LLM-based judgment, and ANLS fallback. Text fields combine semantic similarity with normalized edit similarity. List fields use delimiter-normalized Set-F1 over canonicalized entries. Numeric fields combine numerical overlap with character-level similarity. All field-level scores are normalized to $[0,1]$ before calibration.

\paragraph{Structured output parsing.}
All models are prompted to produce a final answer and a structured dictionary over the predefined reasoning-dimension pool. Missing fields are treated as empty predictions. Fields outside the predefined pool are ignored. If a field contains multiple candidates, delimiters are normalized before scoring. If an output cannot be parsed after deterministic recovery, the corresponding field receives a zero score unless it is removed by the uncertainty or missing-reference calibration rule.

\subsection{Task-Adaptive Weight Generation and Validation}
\label{appendix:reascore_weight_generation}

For each of the 24 reasoning-process tasks in \Reasonbenchmark{}, we generate one task-specific process-field weight vector and cache it for all model evaluations. Concretely, we use DeepSeek-V4-Pro (\texttt{api.deepseek.com}) with temperature 0.0 and the \texttt{json\_object} response format. For each task, the judge model receives three inputs: the task description, the domain-specific reasoning-process field pool, and the expert role annotation for each field. The four roles are \textit{target}, \textit{supporting}, \textit{contextual}, and \textit{excluded}. The judge returns a JSON object that assigns a non-negative weight and a short rationale to each non-excluded field and zero weight to excluded fields.

The output is accepted only if it satisfies three constraints:
\begin{equation}
\left\{
\begin{aligned}
w^{(t)}_j &\ge 0, \quad j\in\mathcal{P}_d, \\
\sum_{j\in\mathcal{P}_d} w^{(t)}_j &= 1, \\
w^{(t)}_j &= 0 \quad \text{if } j \text{ is excluded}.
\end{aligned}
\right.
\end{equation}

The weights are cached per task and released with the benchmark, together with the prompt template, the system prompt, the generated JSON weights, the validation logs, and the cached weight files per task. This validation makes the adaptive weighting reproducible at evaluation time: all evaluated models are scored with the same cached weight vector for a given task.

The adaptive weights capture within-role differences that a fixed role-level rule cannot express. For example, in a Calligraphy author-identification task, \textit{author} is the target field, but \textit{seal text}, \textit{inscription}, and \textit{script style} may receive different supporting weights because they have different diagnostic value. In a Painting technique-identification task, \textit{technique} and \textit{subject} may share more weight, while broader contextual fields receive less. The empirical comparison between LLM-derived weights, human expert judgments, and fixed weighting schemes is reported in the experimental analysis.

\subsection{Uncertainty Calibration}
\label{appendix:reascore_uncertainty}

We treat explicit uncertainty as a distinct output state rather than forcing it into the same category as an incorrect concrete prediction. The uncertainty detector $u(\cdot)$ is triggered by expressions such as ``unknown'', ``uncertain'', ``cannot be determined'', and their Chinese equivalents. When the reference is available, explicit uncertainty receives limited credit because the model has failed to recover the required field. When the reference is genuinely unavailable, explicit uncertainty receives higher partial credit because it avoids hallucination. A concrete prediction for a missing reference is excluded from the scoring denominator rather than being rewarded. This calibration is applied before aggregation.

The constants 0.30 and 0.50 are conservative calibration values. The value 0.30 gives limited credit to explicit uncertainty when the reference exists, while 0.50 gives higher partial credit when the reference is genuinely unavailable. These values are designed to distinguish honest uncertainty from both incorrect concrete predictions and unsupported specificity.

We also track missing-reference cases and unsupported concrete predictions. Unsupported specificity refers to cases where the reference value is unavailable but the model produces a concrete non-empty prediction. This diagnostic helps quantify whether a model tends to make unverifiable claims under incomplete metadata.

\section{Model Suite and Deployment}
\label{appendix:model_details}

\paragraph{Evaluated models and deployment.}
As summarized in Table~\ref{tab:model_details}, we evaluate a representative suite of recent vision-language models, covering both proprietary API-based systems and open-source models with publicly available weights. 
The proprietary models include GPT-5.5\footnote{\url{https://developers.openai.com/api/docs/models/gpt-5.5}}, GPT-5.4-mini\footnote{\url{https://developers.openai.com/api/docs/models/gpt-5.4-mini}}, Gemini-3-Flash-Preview\footnote{\url{https://aistudio.google.com/models/gemini-3}}, and Gemini-2.5-Flash\footnote{\url{https://ai.google.dev/gemini-api/docs/models}}, whose parameter scales are not publicly disclosed and which are evaluated through their official APIs. 
The open-source group includes GLM-4.5V~\cite{glm45v}, Qwen3-VL-8B-Instruct and Qwen3-VL-32B-Instruct~\cite{qw3vl}, Qwen3.5-9B\footnote{\url{https://huggingface.co/Qwen/Qwen3.5-9B}} and Qwen3.5-27B\footnote{\url{https://huggingface.co/Qwen/Qwen3.5-27B}}, as well as the InternVL3 and InternVL3.5 series~\cite{internvl3,internvl35}. 
For locally deployed open-source models, we use SGLang with BF16 precision to ensure a consistent inference backend, while API-only models are evaluated through their official endpoints.

\paragraph{Auxiliary models.}
In addition to the evaluated vision-language models, we use two text-only DeepSeek-V4 models for auxiliary components in the evaluation pipeline. 
DeepSeek-V4-Flash is used as the auxiliary answer judge, whereas DeepSeek-V4-Pro is used only for \textsc{ReaScore} weight generation, where task-specific weights are derived from task descriptions, reasoning-dimension definitions, and expert-role annotations.\footnote{\url{https://api-docs.deepseek.com/news/news260424}} 
Since these two models do not process multimodal inputs in our benchmark and serve only judging or weighting purposes, they are excluded from the main set of competing models.

\begin{table*}[t]
\centering
\small
\setlength{\tabcolsep}{3.0pt}
\renewcommand{\arraystretch}{1.08}
\resizebox{\textwidth}{!}{
\begin{tabular}{@{}l *{6}{c}@{}}
\toprule
\textbf{Model} & \textbf{Open-Source} & \textbf{Parameters} & \textbf{Institution} & \textbf{Deployment} & \textbf{Modality} & \textbf{Role} \\
\midrule
GPT-5.5 & No & -- & OpenAI & Official API & VL & Evaluated \\
GPT-5.4-mini & No & -- & OpenAI & Official API & VL & Evaluated \\
Gemini-3-Flash-Preview & No & -- & Google & Official API & VL & Evaluated \\
Gemini-2.5-Flash & No & -- & Google & Official API & VL & Evaluated \\
GLM-4.5V & Yes & 12B (MoE) & Z.ai & Official API & VL & Evaluated \\
Qwen3-VL-8B-Instruct & Yes & 8B & Alibaba & \mbox{SGLang, BF16} & VL & Evaluated \\
\mbox{Qwen3-VL-32B-Instruct} & Yes & 32B & Alibaba & \mbox{SGLang, BF16} & VL & Evaluated \\
Qwen3.5-9B & Yes & 9B & Alibaba & \mbox{SGLang, BF16} & VL & Evaluated \\
Qwen3.5-27B & Yes & 27B & Alibaba & \mbox{SGLang, BF16} & VL & Evaluated \\
InternVL3-8B & Yes & 8B & \mbox{Shanghai AI Lab} & \mbox{SGLang, BF16} & VL & Evaluated \\
InternVL3-14B & Yes & 14B & \mbox{Shanghai AI Lab} & \mbox{SGLang, BF16} & VL & Evaluated \\
InternVL3.5-8B-Instruct & Yes & 8B & \mbox{Shanghai AI Lab} & \mbox{SGLang, BF16} & VL & Evaluated \\
InternVL3.5-14B-Instruct & Yes & 14B & \mbox{Shanghai AI Lab} & \mbox{SGLang, BF16} & VL & Evaluated \\
InternVL3.5-30B-A3B-Instruct & Yes & 3B (MoE) & \mbox{Shanghai AI Lab} & \mbox{SGLang, BF16} & VL & Evaluated \\
DeepSeek-V4-Flash & Yes & 13B (MoE) & DeepSeek-AI & Official API & Text & Auxiliary answer judge \\
DeepSeek-V4-Pro & Yes & 49B (MoE) & DeepSeek-AI & Official API & Text & \textsc{ReaScore} weight generation \\
\bottomrule
\end{tabular}
}
\caption{Details of evaluated and judge models.}
\label{tab:model_details}
\end{table*}

\section{Experimental Details}
\label{appendix:experimental_details}

\subsection{Prompting, Decoding, and Output Normalisation}
\label{appendix:prompting_decoding_normalisation}

All evaluated models receive identical task-formatted prompts and the same compressed image inputs. Image compression follows the data postprocessing protocol in Appendix~\ref{appendix:postprocessing}. Decoding is deterministic, with temperature approximately 0, and each model generates one response per instance. No external retrieval, web search, or tool use is allowed.

For \benchmark{}, prompts require only the task answer, such as an option letter, a target span, or a concise free-form response. For \Reasonbenchmark{}, prompts require a structured output containing the domain-level reasoning-dimension pool and the final answer. Output normalisation includes option-letter extraction, span trimming, delimiter normalisation for list-valued dimensions, uncertainty-token detection, and task-specific answer parsing.

\subsection{Scoring, Judging, and Caching}
\label{appendix:experimental_scoring_caching}

The same task-specific answer scorer is used for answer-only \benchmark{} runs and structured \Reasonbenchmark{} runs, so the final-answer score remains directly comparable across settings. Task-level metrics and aggregation rules are provided in Appendix~\ref{appendix:metrics}, with the full task-to-metric mapping in Table~\ref{tab:task_specific_metrics}.

For long-form QA and translation tasks, \textsc{CulMind-Score} uses DeepSeek-V4-Flash as an auxiliary LLM judge. The judge rates accuracy, completeness, relevance, and fluency on a 0--10 scale, and the resulting score is normalised to $[0,1]$. We sample up to 50 instances per task and model for this auxiliary judging component.

For \Reasonbenchmark{}, model outputs are frozen before scoring. The role-uniform variant and \textsc{ReaScore} use the same outputs, answer scorers, type-aware dimension scorers, and uncertainty calibration; they differ only in the dimension weights. \textsc{ReaScore} weights are generated once per task, checked for non-negativity, sum-to-one normalisation, and zero mass on \textit{excluded} dimensions, then cached for all model evaluations. Fixed-weight and task-adaptive scores are computed from the same frozen model outputs, so differences between the two settings reflect the scoring scheme rather than additional model generations.

\section{Evaluation Metrics}
\label{appendix:metrics}

Table~\ref{tab:task_specific_metrics} lists the primary and auxiliary evaluation metrics for all 50 \benchmark{} tasks.

\begin{table*}[!htbp]
\centering
\small
\setlength{\tabcolsep}{3.0pt}
\renewcommand{\arraystretch}{1.04}
\begin{tabularx}{\textwidth}{@{}l >{\raggedright\arraybackslash}p{0.29\textwidth} l >{\raggedright\arraybackslash}X@{}}
\toprule
\textbf{Task} & \textbf{Task name} & \textbf{Primary metric} & \textbf{Auxiliary metrics} \\
\midrule

\rowcolor{gray!15}
\multicolumn{4}{@{}l}{\textbf{Ancient Books}} \\
T1  & Text OCR & CR & AR, NormEdit, \mbox{OCR-F1}, \mbox{OCR-Prec.}, \mbox{OCR-Rec.}, \mbox{OCR-BLEU} \\
T2  & Source Attribution & Acc & -- \\
T3  & Author Identification & Acc & -- \\
T4  & Dynasty Identification & Acc & -- \\
T5  & Version Identification & Acc & -- \\
T6  & Punctuation and Segmentation & Punctuation-F1 & Punctuation precision, punctuation recall \\
T7  & Structural Organization & Tag-F1 & Tag precision, tag recall, tag-type Acc, AR, CR \\
T8  & Named Entity Recognition & NER-F1 & NER precision, NER recall \\
T9  & Word Explanation & Acc & -- \\
T10 & Classical Chinese to Modern Chinese & BLEU & \mbox{ROUGE-1}, \mbox{ROUGE-2}, \mbox{ROUGE-L} \\
T11 & Reading Comprehension & BERTScore & ANLS, \mbox{LLM-quality} \\

\rowcolor{gray!15}
\multicolumn{4}{@{}l}{\textbf{Calligraphy}} \\
T12 & Calligraphy OCR & CR & AR, NormEdit, \mbox{OCR-F1}, \mbox{OCR-Prec.}, \mbox{OCR-Rec.}, \mbox{OCR-BLEU} \\
T13 & Title Identification & Acc & -- \\
T14 & Author Identification & Acc & -- \\
T15 & Dynasty Identification & Acc & -- \\
T16 & Script Identification & Acc & -- \\
T17 & Form Identification & Acc & -- \\
T18 & Collection Identification & Acc & -- \\
T19 & Content Description & Set-F1 & Set-EM \\
T20 & Style/School Identification & Acc & -- \\
T21 & Calligraphy Appreciation & BERTScore & ANLS, \mbox{LLM-quality} \\

\rowcolor{gray!15}
\multicolumn{4}{@{}l}{\textbf{Painting}} \\
T22 & Painting Seal Recognition & CR & AR, NormEdit, \mbox{OCR-F1}, \mbox{OCR-Prec.}, \mbox{OCR-Rec.}, \mbox{OCR-BLEU} \\
T23 & Title Identification & Acc & -- \\
T24 & Author Identification & Acc & -- \\
T25 & Dynasty Identification & Acc & -- \\
T26 & Collection Identification & Acc & -- \\
T27 & Form Identification & Acc & -- \\
T28 & Technique Identification & Acc & -- \\
T29 & Visual Content Recognition & Set-F1 & Set-EM \\
T30 & Painting Appreciation & BERTScore & ANLS, \mbox{LLM-quality} \\

\rowcolor{gray!15}
\multicolumn{4}{@{}l}{\textbf{Seals}} \\
T31 & Seal OCR & CR & AR, NormEdit, \mbox{OCR-F1}, \mbox{OCR-Prec.}, \mbox{OCR-Rec.}, \mbox{OCR-BLEU} \\
T32 & Author Identification & Acc & -- \\
T33 & Dynasty Identification & Acc & -- \\
T34 & Source Attribution & Acc & -- \\
T35 & Seal Type Identification & Acc & -- \\
T36 & Shape Identification & Acc & -- \\
T37 & Usage Identification & Acc & -- \\

\rowcolor{gray!15}
\multicolumn{4}{@{}l}{\textbf{Decorative Arts}} \\
T38 & Name Identification & Acc & -- \\
T39 & Dynasty Identification & Acc & -- \\
T40 & Visual Question Answering & Acc & -- \\
T41 & Material Identification & Acc & -- \\
T42 & Function Identification & Acc & -- \\

\rowcolor{gray!15}
\multicolumn{4}{@{}l}{\textbf{Cave Murals}} \\
T43 & Source Attribution & Acc & -- \\
T44 & Dynasty Identification & Acc & -- \\
T45 & Theme Identification & Set-F1 & Set-EM \\
T46 & Scene Description & BERTScore & ANLS, \mbox{LLM-quality} \\

\rowcolor{gray!15}
\multicolumn{4}{@{}l}{\textbf{Ancient Scripts}} \\
T47 & Ancient Script OCR & Char-Acc & Char-ANLS \\
T48 & Script Evolution Matching & Set-F1 & Set-EM \\
T49 & Dongba Script OCR & Char-Acc & Char-ANLS \\
T50 & Ancient Script Paragraph Translation & BLEU & \mbox{ROUGE-1}, \mbox{ROUGE-2}, \mbox{ROUGE-L}, BERTScore, ANLS, \mbox{LLM-quality} \\

\bottomrule
\end{tabularx}
\caption{Primary and auxiliary evaluation metrics for the 50 \benchmark{} tasks.}
\label{tab:task_specific_metrics}
\end{table*}

\section{Process-Diagnostic Computation and Threshold Sensitivity}
\label{appendix:reascore_threshold_sensitivity}

The continuous \textsc{ReaScore} does not require thresholding. However, the diagnostic columns in Table~\ref{tab:reasonbenchmark_main} require a binary process-reliability decision. For model \(m\), instance \(i\), and task \(t\), we define
\begin{equation}
p_{i,t}^{(m)}(\theta_R)
=
\mathbf{1}\{R_{i,t}^{(m)} \ge \theta_R\},
\end{equation}
where \(R_{i,t}^{(m)}\) is the instance-level \textsc{ReaScore}. Unless otherwise specified, we use \(\theta_R=0.60\).

Let \(c_{i,t}^{(m)}\in\{0,1\}\) denote the final-answer correctness label produced by the official task-specific answer scorer. We further define \(b_{i,t}^{(m)}\in\{0,1\}\) as a valid-process indicator. It is 1 when the structured reasoning output can be parsed and at least one active reasoning dimension remains after uncertainty and missing-reference calibration. Otherwise, it is 0.

For each model, we compute the answer-conditioned process-reliable diagnostic as
\begin{equation}
\resizebox{0.92\columnwidth}{!}{$
\mathrm{ACPR}^{(m)}(\theta_R)
=
\frac{1}{|\mathcal{T}_m^+|}
\sum_{t\in\mathcal{T}_m^+}
\frac{
\sum_i c_{i,t}^{(m)} b_{i,t}^{(m)} p_{i,t}^{(m)}(\theta_R)
}{
\sum_i c_{i,t}^{(m)}
}
$}
\end{equation}
and the answer-conditioned process-weak diagnostic as
\begin{equation}
\resizebox{0.92\columnwidth}{!}{$
\mathrm{ACPW}^{(m)}(\theta_R)
=
\frac{1}{|\mathcal{T}_m^+|}
\sum_{t\in\mathcal{T}_m^+}
\frac{
\sum_i c_{i,t}^{(m)} b_{i,t}^{(m)} \bigl(1-p_{i,t}^{(m)}(\theta_R)\bigr)
}{
\sum_i c_{i,t}^{(m)}
}
$}
\end{equation}
where \(\mathcal{T}_m^+\) denotes tasks for which model \(m\) has at least one answer-correct instance.

Thus, \(\mathrm{ACPR}\) corresponds to the \textbf{Ans. Correct / Proc. Reliable} column in Table~\ref{tab:reasonbenchmark_main}. It measures the task-macro fraction of answer-correct cases whose valid reasoning process is also reliable. \(\mathrm{ACPW}\) corresponds to the \textbf{Ans. Correct / Proc. Weak} column and measures the corresponding weak-process rate among answer-correct cases with valid process scores. Because invalid or fully removed structured outputs are not counted in either numerator, the two diagnostic columns are not forced to sum to 1 in the macro-averaged table.

\paragraph{Threshold status.}
We use \(\theta_R=0.60\) as a transparent diagnostic cutoff for reporting binary process-reliability columns. The continuous \textsc{ReaScore} remains the primary process-level metric and does not require thresholding. The expert study in Appendix~\ref{appendix:expert_annotation} is used to validate whether \textsc{ReaScore} aligns with human judgments of reasoning-process reliability, rather than to tune this threshold.

\paragraph{Sensitivity protocol.}
To verify that the diagnostic conclusion does not depend on a single cutoff, we vary \(\theta_R\) over a deterministic grid around the default value and also examine stochastic perturbations within the same operating range. The purpose of this analysis is to test whether threshold changes alter the main finding that correct final answers often have weak reasoning processes. In our evaluation, changing the threshold affects the absolute process-pass rate, but does not change the qualitative conclusion that reliable answer-process alignment remains rare.

\section{Expert Annotation Protocol}
\label{appendix:expert_annotation}

We use expert annotation to provide an external reference for evaluating reasoning-process reliability. The annotation set contains 240 model-instance outputs from Gemini-3-Flash-Preview, sampled as 10 instances from each of the 24 reasoning tasks. Each output is independently rated by three CCH experts. Annotators are shown the image, question, model final answer, model-generated reasoning dimensions, and reference reasoning dimensions. Model identities and automatic scores are hidden.

Table~\ref{tab:expert_annotation_dimensions} reports the four annotation dimensions.

\begin{table}[H]
\centering
\small
\setlength{\tabcolsep}{3.5pt}
\renewcommand{\arraystretch}{1.03}
\resizebox{\columnwidth}{!}{
\begin{tabular}{p{0.32\columnwidth}p{0.62\columnwidth}}
\toprule
\textbf{Dimension} & \textbf{Annotation Criterion} \\
\midrule
Dimension Correctness &
Whether the predicted intermediate dimensions are factually correct. A score of 5 means all dimensions are correct, while a score of 1 means most dimensions are incorrect. \\

Evidence Coverage &
Whether the target and supporting dimensions required by the task are adequately covered. A score of 5 means all key evidence is present, while a score of 1 means key evidence is missing. \\

Inference Sufficiency &
Whether the intermediate dimensions sufficiently support the final answer. A score of 5 means the evidence fully supports the answer, while a score of 1 means the evidence is irrelevant or contradictory. \\

Uncertainty Calibration &
Whether the output avoids unsupported specificity. A score of 5 means uncertainty is properly calibrated, while a score of 1 indicates overconfident or hallucinated claims. \\
\bottomrule
\end{tabular}
}
\caption{
Expert annotation dimensions for reasoning-process reliability. Each dimension is rated on a 1 to 5 Likert scale.
}
\label{tab:expert_annotation_dimensions}
\end{table}

The expert process score is computed as the mean of the four annotation dimensions. We also derive a binary expert-pass label for AUROC and F1 evaluation. An output is marked as expert-pass when its expert process score is at least 4.0 and it contains no fatal error. Fatal errors include wrong target dimensions, hallucinated key evidence, or contradictions with visible image content.

Table~\ref{tab:expert_dimension_statistics} reports per-dimension annotation statistics.

\begin{table}[H]
\centering
\small
\setlength{\tabcolsep}{5pt}
\renewcommand{\arraystretch}{1.08}
\resizebox{\columnwidth}{!}{
\begin{tabular}{lccccc}
\toprule
\textbf{Dimension} & \textbf{Mean} & \textbf{Std.} & \textbf{Min} & \textbf{Median} & \textbf{Max} \\
\midrule
Dimension Correctness & 3.41 & 1.17 & 1 & 3 & 5 \\
Evidence Coverage & 3.99 & 0.59 & 2 & 4 & 5 \\
Inference Sufficiency & 3.40 & 1.28 & 1 & 3 & 5 \\
Uncertainty Calibration & 2.83 & 0.65 & 1 & 3 & 5 \\
\bottomrule
\end{tabular}
}
\caption{Per-dimension statistics of expert annotations.}
\label{tab:expert_dimension_statistics}
\end{table}

Table~\ref{tab:expert_inter_annotator_reliability} reports inter-annotator reliability.

\begin{table}[H]
\centering
\small
\setlength{\tabcolsep}{5pt}
\renewcommand{\arraystretch}{1.08}
\resizebox{\columnwidth}{!}{
\begin{tabular}{lc p{0.5\columnwidth}}
\toprule
\textbf{Metric} & \textbf{Value} & \textbf{Meaning} \\
\midrule
ICC(2,3) & 0.9656 & Agreement on continuous expert process scores across three raters. \\
Fleiss' $\kappa$ & 0.8164 & Multi-rater agreement on binary expert-pass labels, corrected for chance. \\
Gwet's AC1 & 0.8732 & Agreement on binary labels with better robustness under skewed label prevalence. \\
\bottomrule
\end{tabular}
}
\caption{Inter-annotator reliability of expert judgments.}
\label{tab:expert_inter_annotator_reliability}
\end{table}

We also report per-dimension means by task group to show how expert ratings differ between surface and integrative reasoning tasks. Table~\ref{tab:expert_dimension_by_group} summarizes these results.

\begin{table}[H]
\centering
\small
\setlength{\tabcolsep}{5pt}
\renewcommand{\arraystretch}{1.08}
\resizebox{\columnwidth}{!}{
\begin{tabular}{lcc}
\toprule
\textbf{Dimension} & \textbf{Surface} & \textbf{Integrative} \\
\midrule
Dimension Correctness & 3.56 & 3.36 \\
Evidence Coverage & 3.77 & 4.06 \\
Inference Sufficiency & 3.32 & 3.43 \\
Uncertainty Calibration & 2.96 & 2.79 \\
\bottomrule
\end{tabular}
}
\caption{Per-dimension expert scores by reasoning group.}
\label{tab:expert_dimension_by_group}
\end{table}
\clearpage
\onecolumn

\section{\benchmark{} Examples}
\label{appendix:benchmark_dataset_examples}

\begin{figure}[!htbp]
    \centering
    \includegraphics[
        width=0.95\textwidth,
        height=0.72\textheight,
        keepaspectratio
    ]{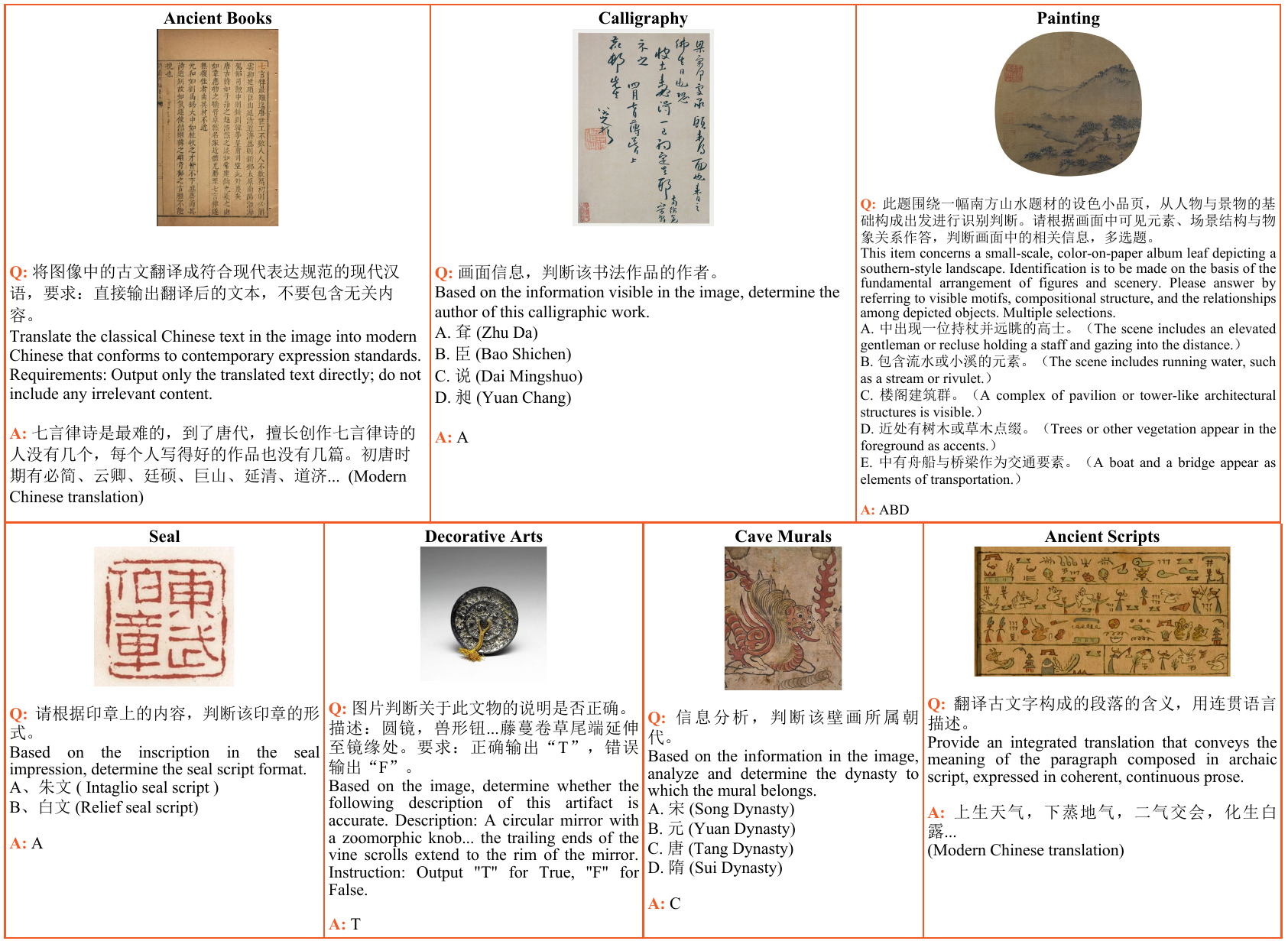}
    \caption{Data examples from \benchmark{}.}
    \label{fig:benchmark_subdomains}
\end{figure}

\begin{figure*}[p]
    \centering
    \includegraphics[width=0.95\textwidth]{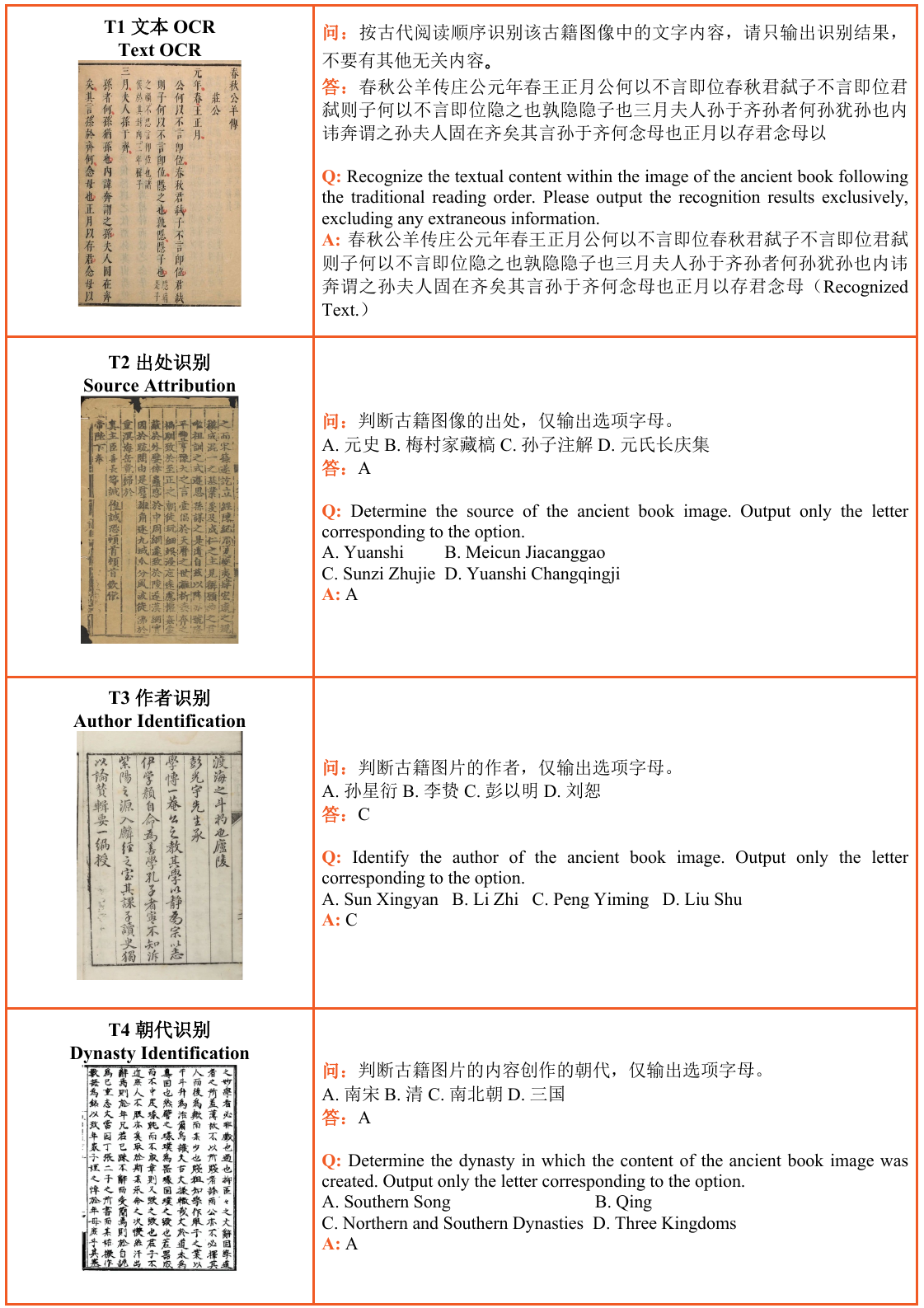}
    \caption{\benchmark{} examples for Tasks 1--4.}
    \label{fig:benchmark_dataset_example_1}
\end{figure*}

\begin{figure*}[p]
    \centering
    \includegraphics[width=0.95\textwidth]{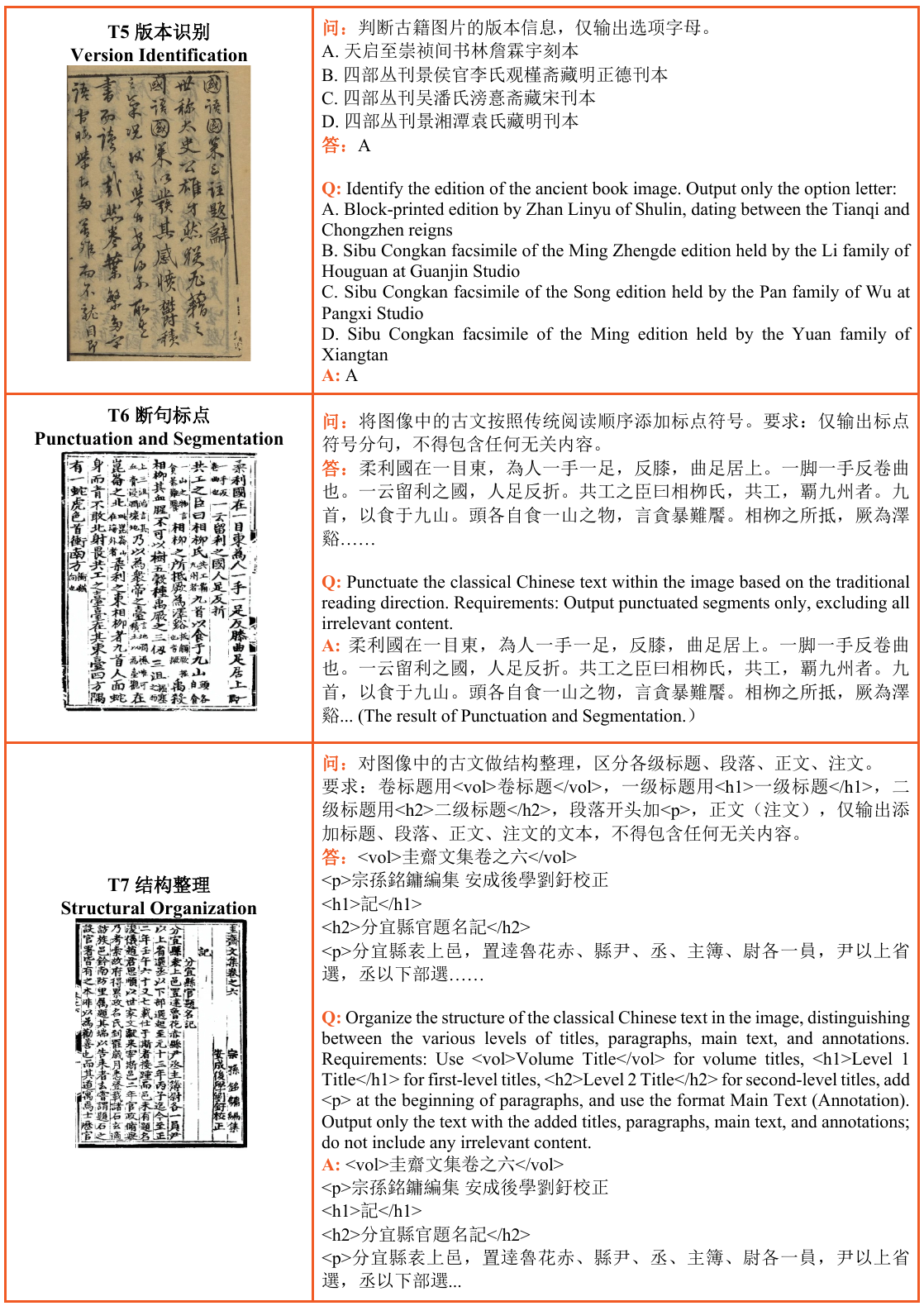}
    \caption{\benchmark{} examples for Tasks 5--7.}
    \label{fig:benchmark_dataset_example_2}
\end{figure*}

\begin{figure*}[p]
    \centering
    \includegraphics[width=0.95\textwidth]{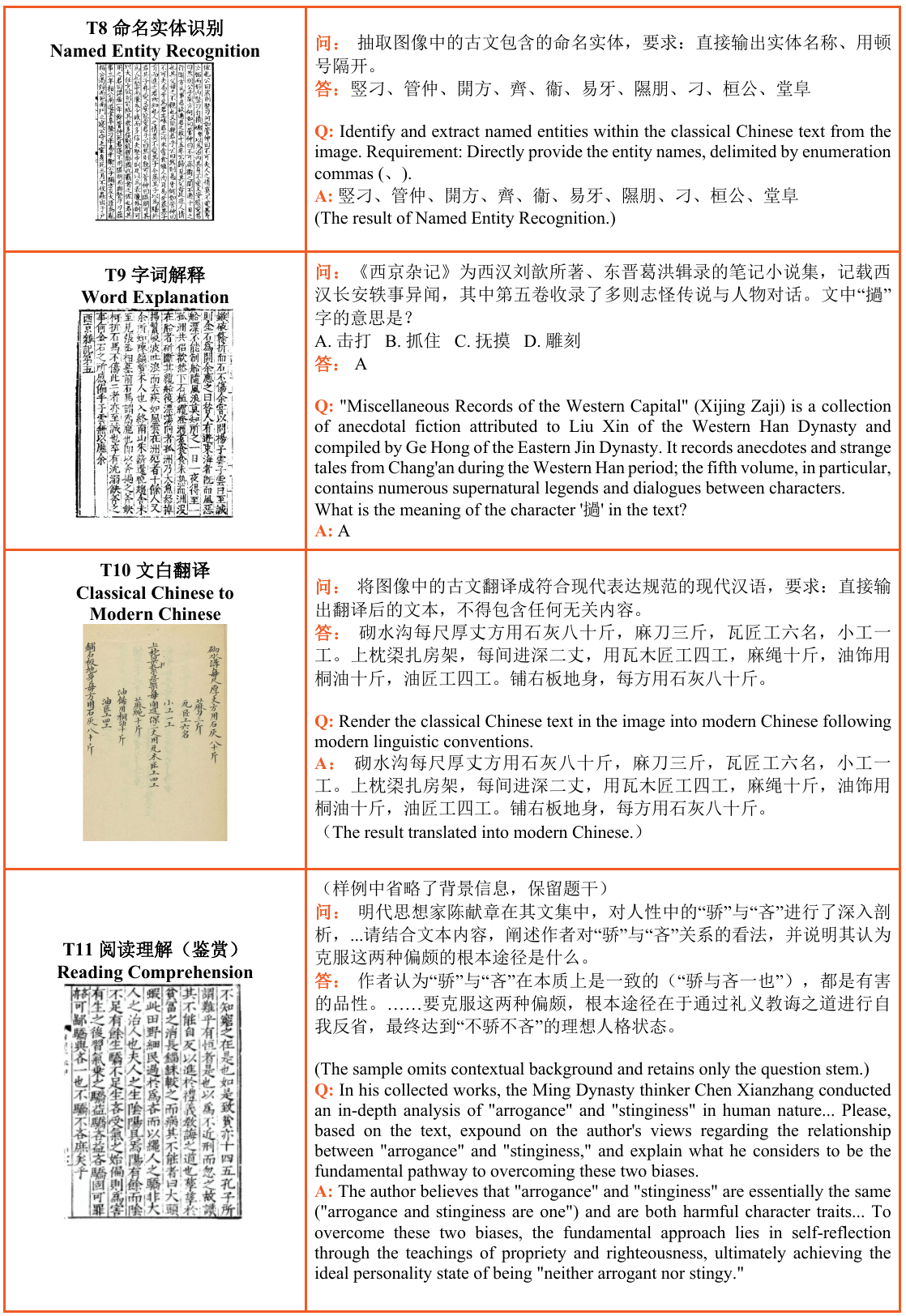}
    \caption{\benchmark{} examples for Tasks 8--11.}
    \label{fig:benchmark_dataset_example_3}
\end{figure*}

\begin{figure*}[p]
    \centering
    \includegraphics[width=0.95\textwidth]{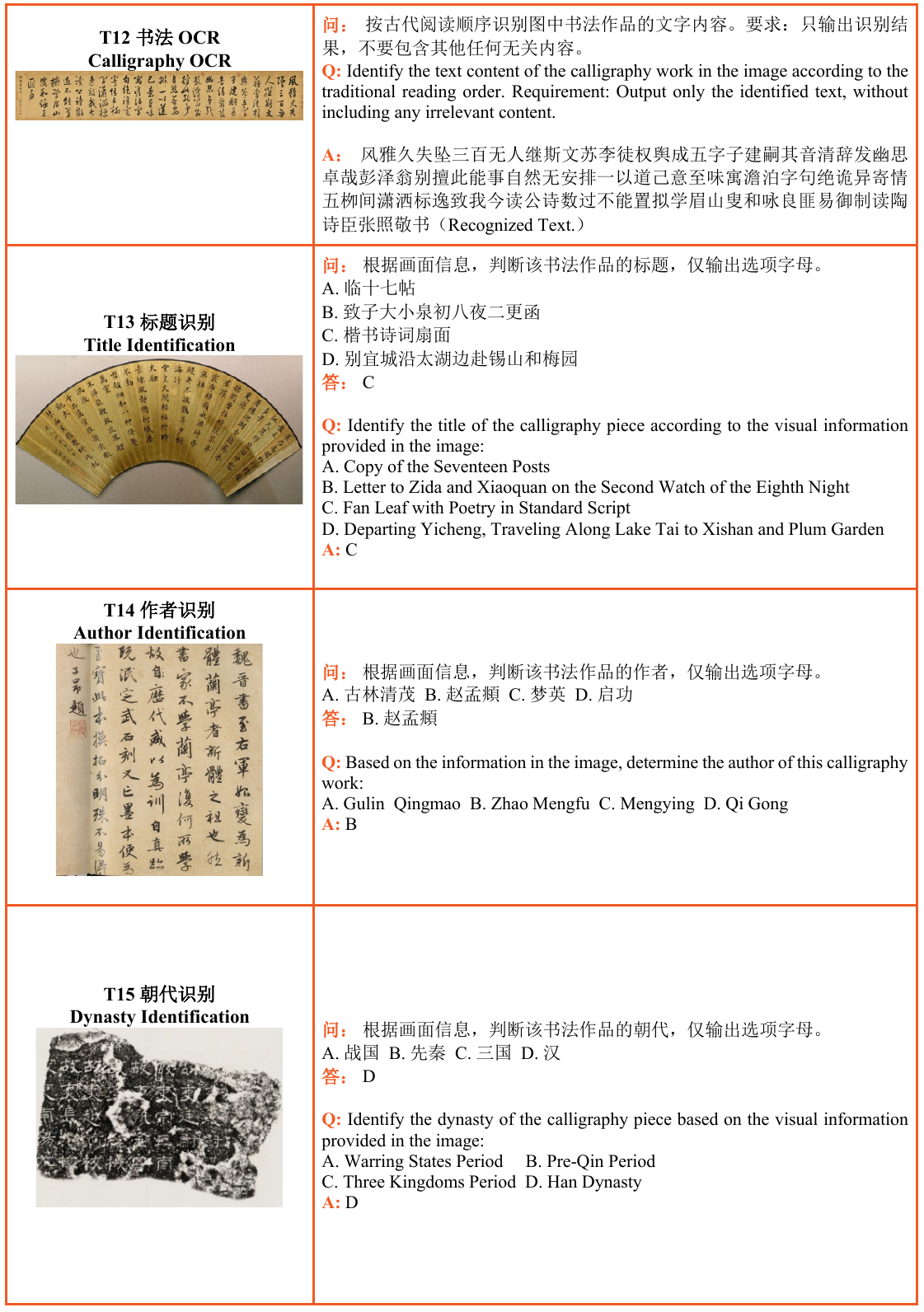}
    \caption{\benchmark{} examples for Tasks 12--15.}
    \label{fig:benchmark_dataset_example_4}
\end{figure*}

\begin{figure*}[p]
    \centering
    \includegraphics[width=0.95\textwidth]{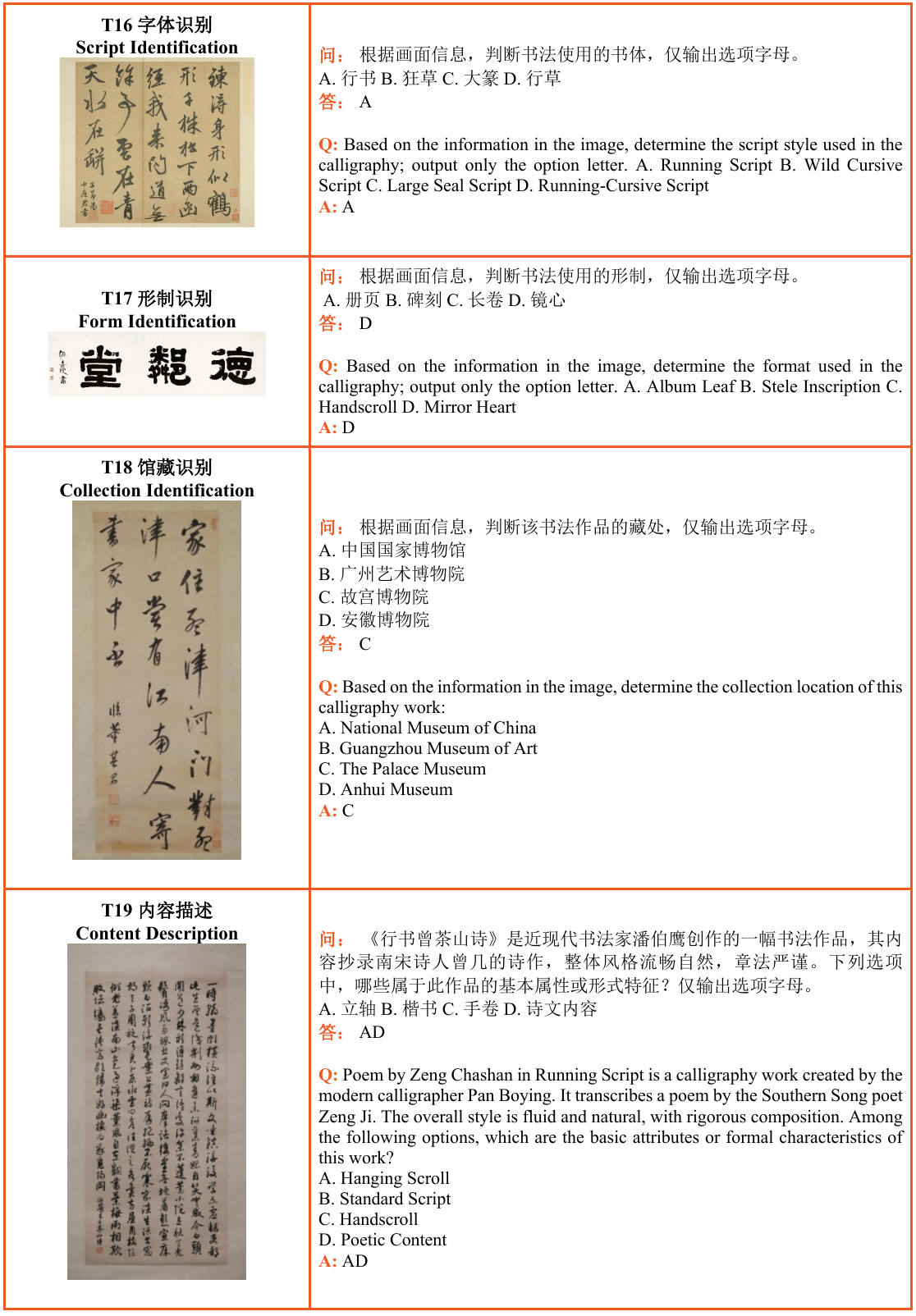}
    \caption{\benchmark{} examples for Tasks 16--19.}
    \label{fig:benchmark_dataset_example_5}
\end{figure*}

\begin{figure*}[p]
    \centering
    \includegraphics[width=0.95\textwidth]{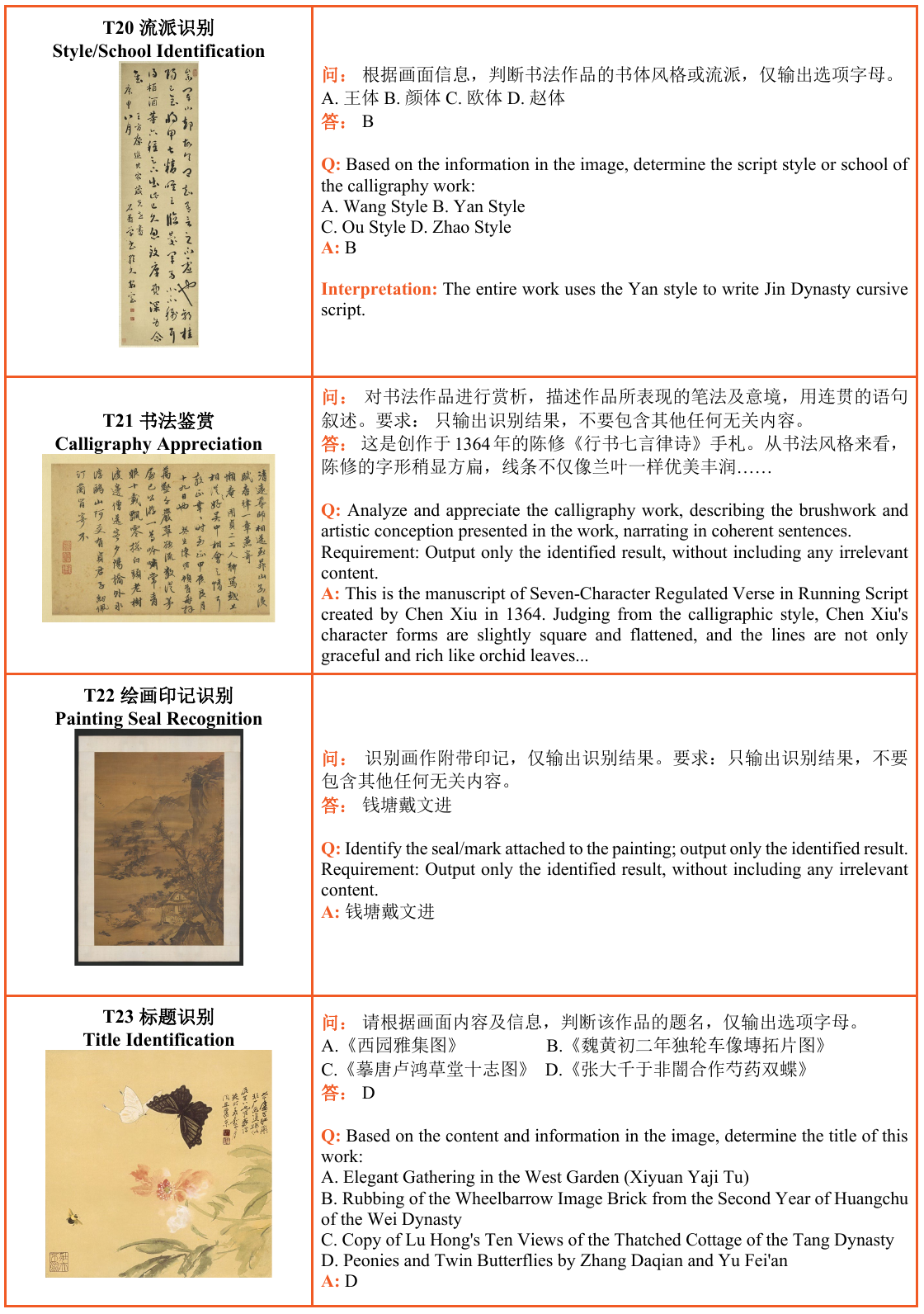}
    \caption{\benchmark{} examples for Tasks 20--23.}
    \label{fig:benchmark_dataset_example_6}
\end{figure*}

\begin{figure*}[p]
    \centering
    \includegraphics[width=0.95\textwidth]{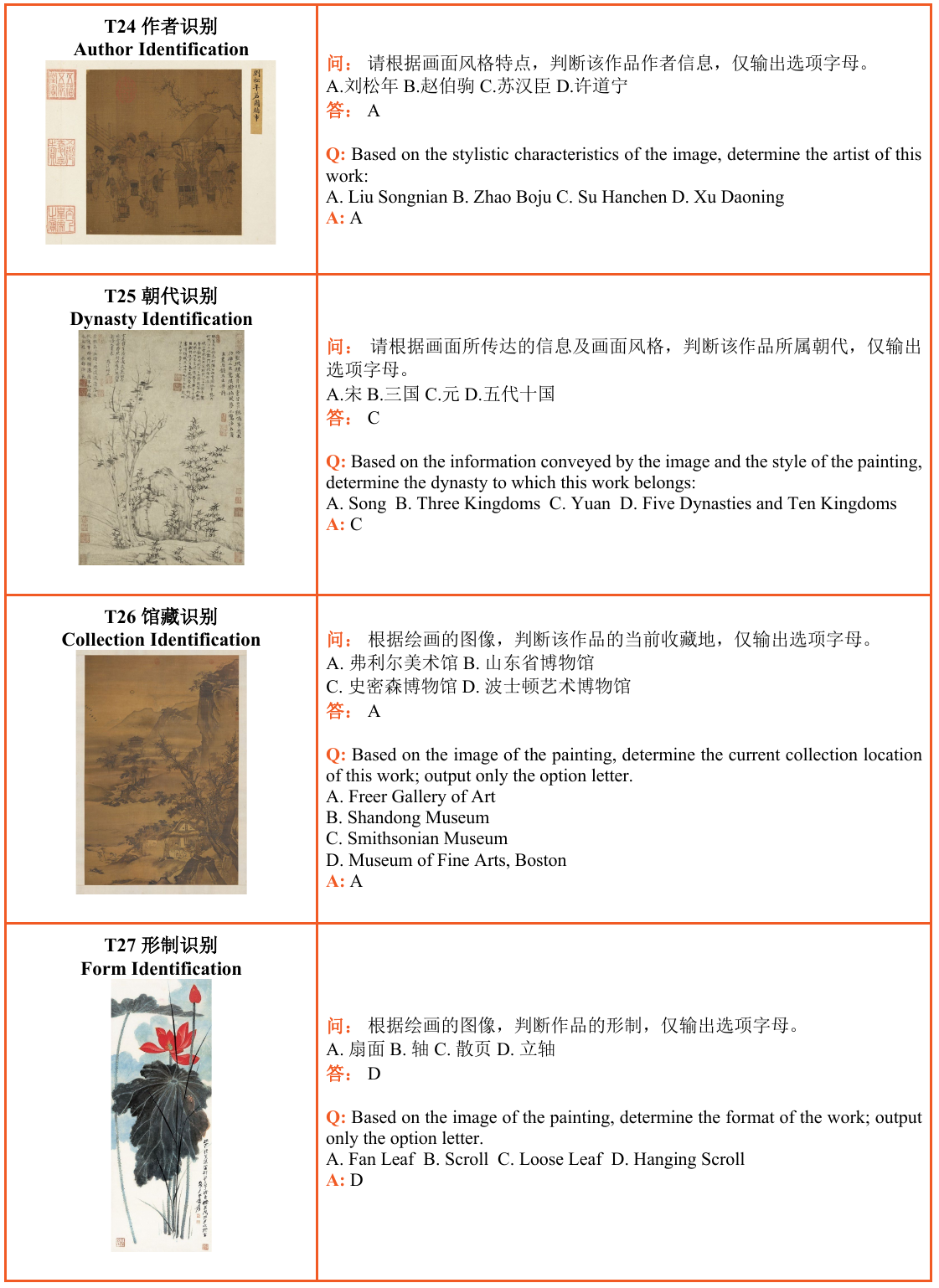}
    \caption{\benchmark{} examples for Tasks 24--27.}
    \label{fig:benchmark_dataset_example_7}
\end{figure*}

\begin{figure*}[p]
    \centering
    \includegraphics[width=0.95\textwidth]{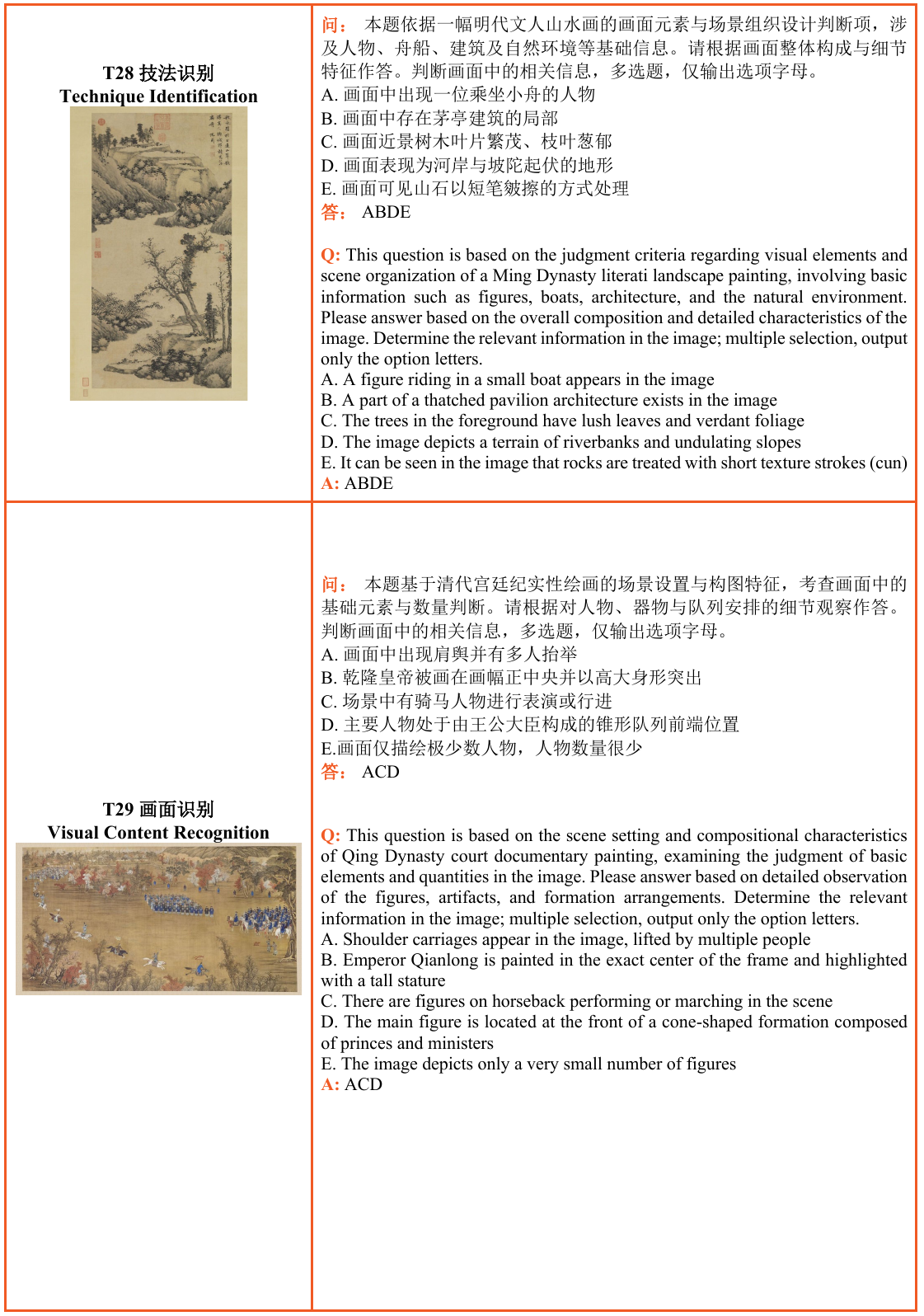}
    \caption{\benchmark{} examples for Tasks 28--29.}
    \label{fig:benchmark_dataset_example_8}
\end{figure*}

\begin{figure*}[p]
    \centering
    \includegraphics[width=0.95\textwidth]{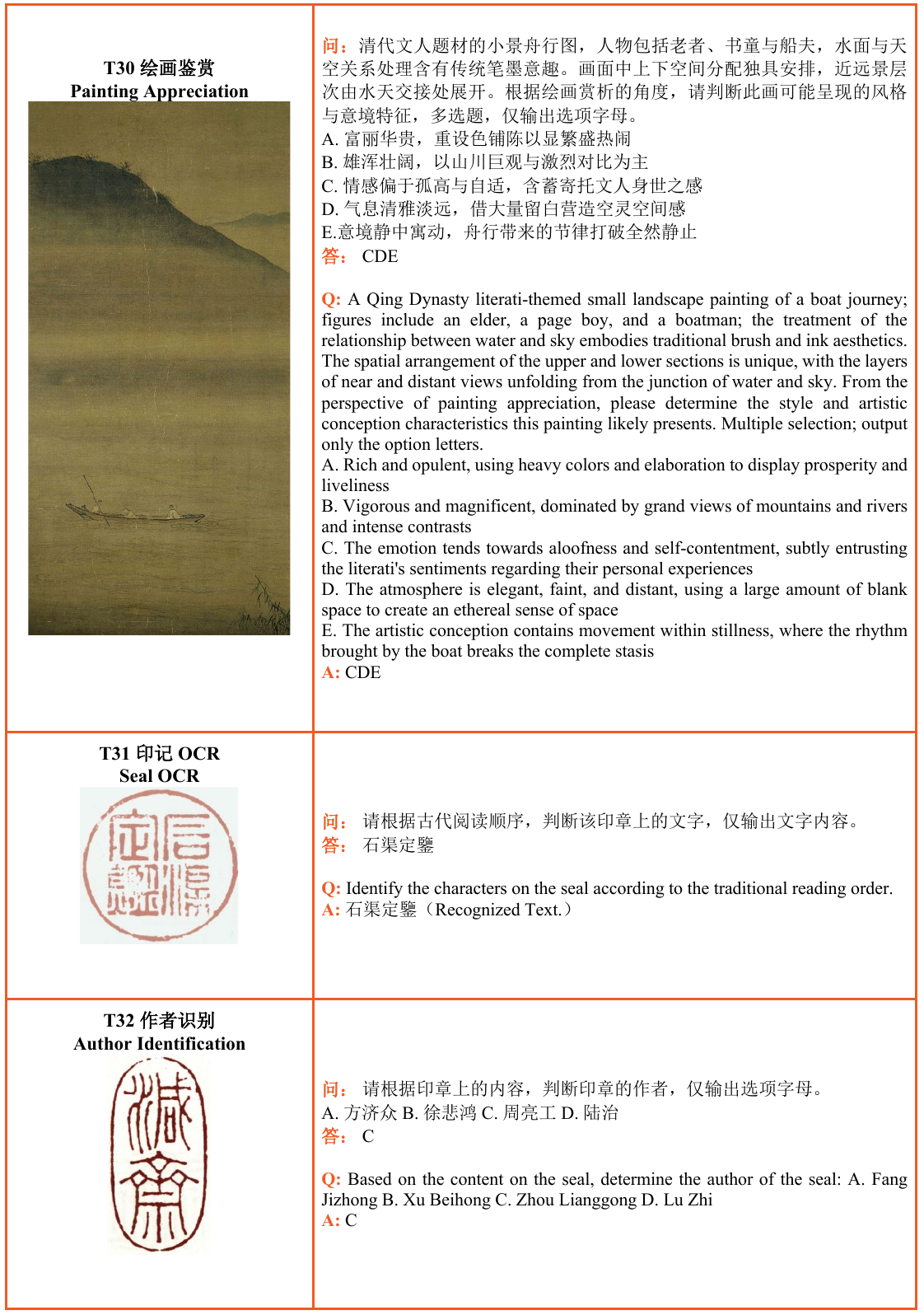}
    \caption{\benchmark{} examples for Tasks 30--32.}
    \label{fig:benchmark_dataset_example_9}
\end{figure*}

\begin{figure*}[p]
    \centering
    \includegraphics[width=0.95\textwidth]{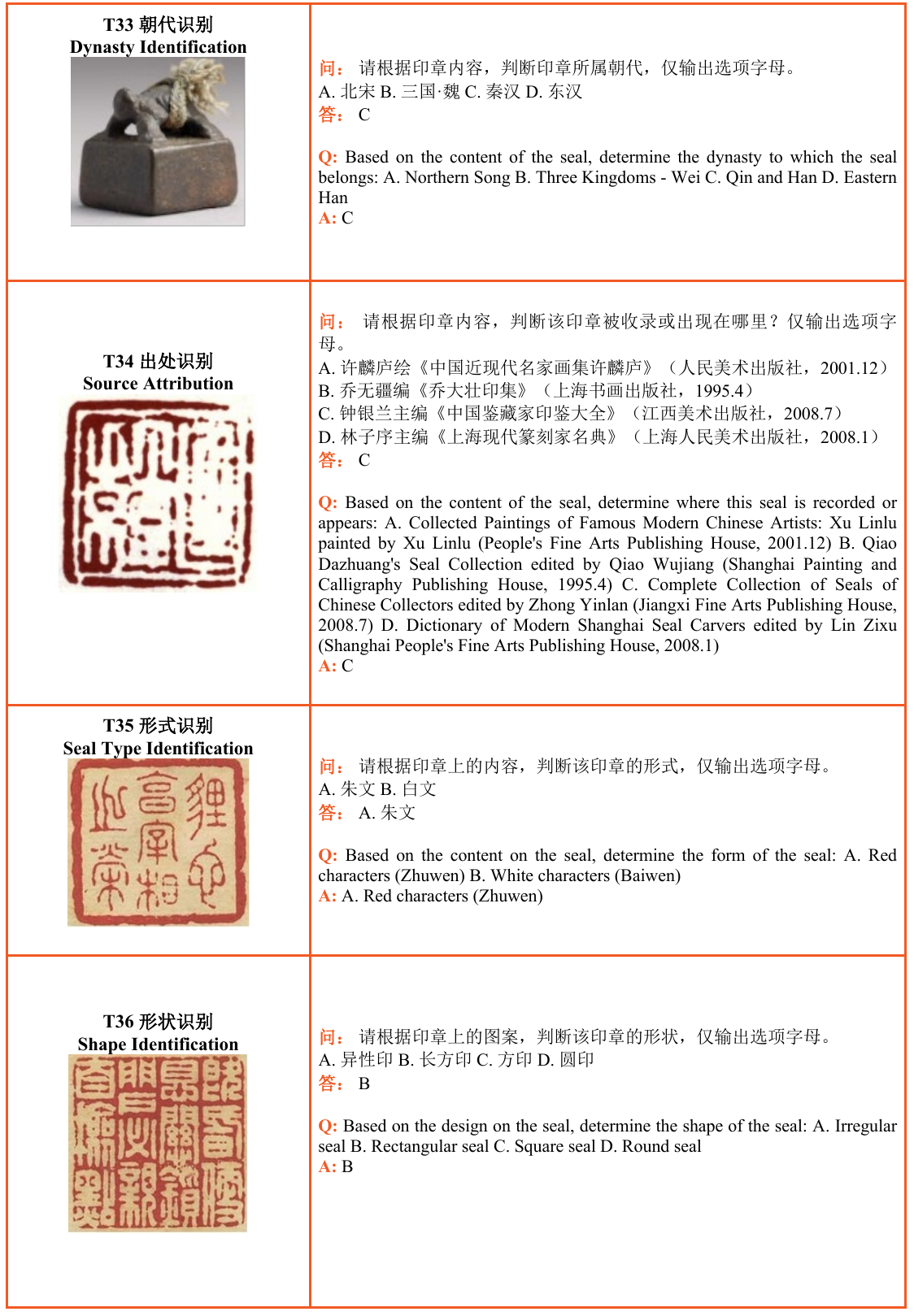}
    \caption{\benchmark{} examples for Tasks 33--36.}
    \label{fig:benchmark_dataset_example_10}
\end{figure*}

\begin{figure*}[p]
    \centering
    \includegraphics[width=0.95\textwidth]{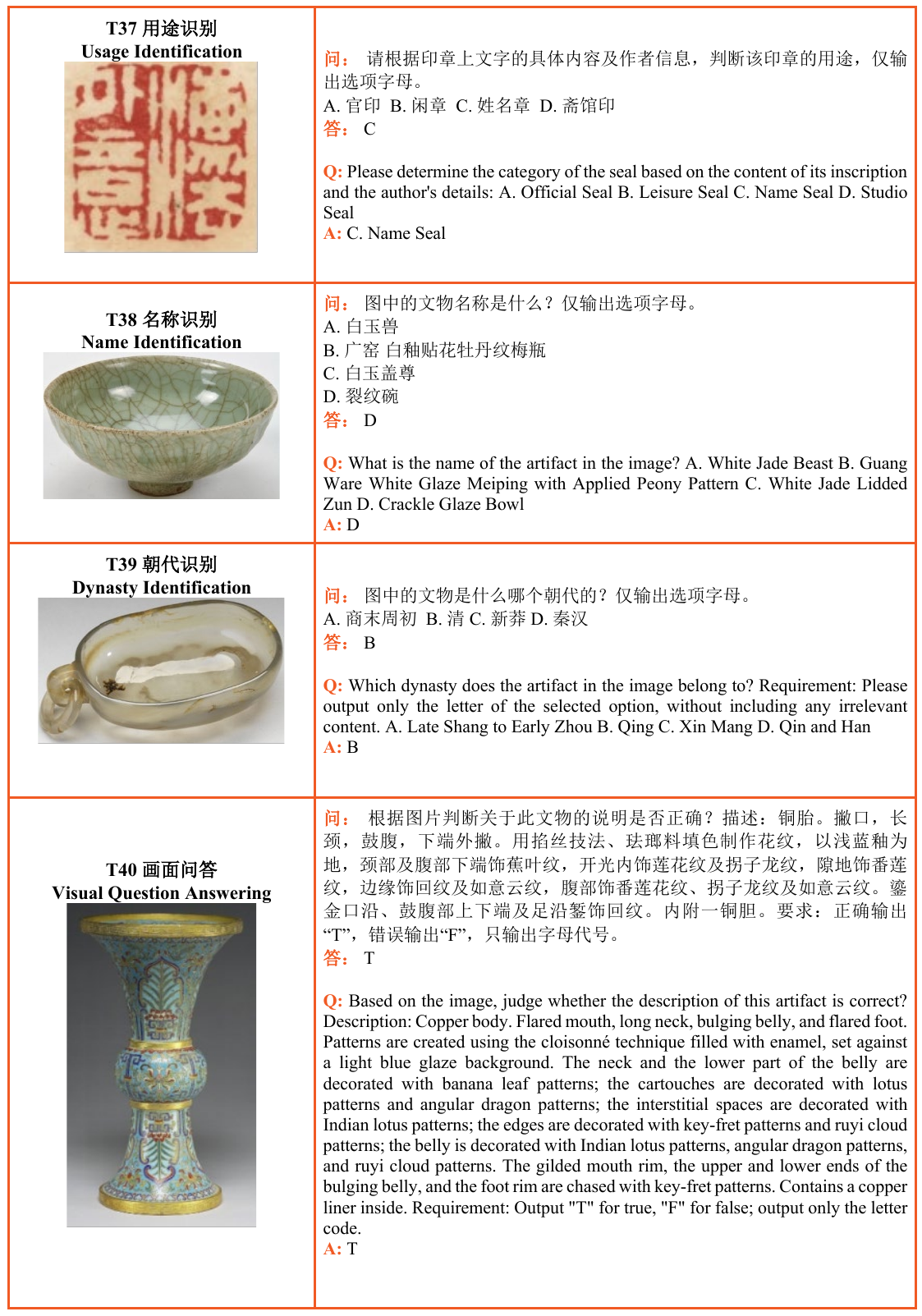}
    \caption{\benchmark{} examples for Tasks 37--40.}
    \label{fig:benchmark_dataset_example_11}
\end{figure*}

\begin{figure*}[p]
    \centering
    \includegraphics[width=0.95\textwidth]{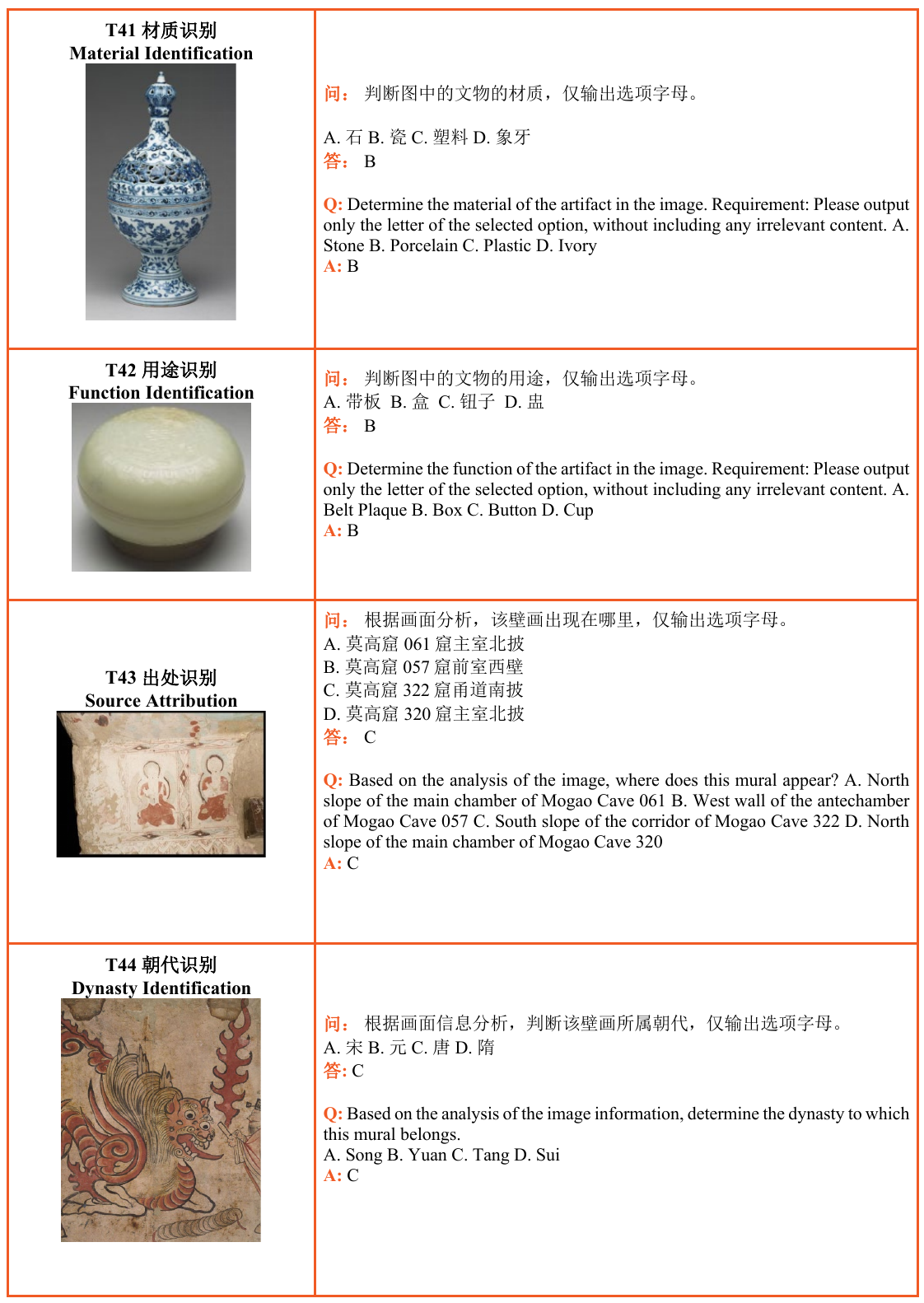}
    \caption{\benchmark{} examples for Tasks 41--44.}
    \label{fig:benchmark_dataset_example_12}
\end{figure*}

\begin{figure*}[p]
    \centering
    \includegraphics[width=0.95\textwidth]{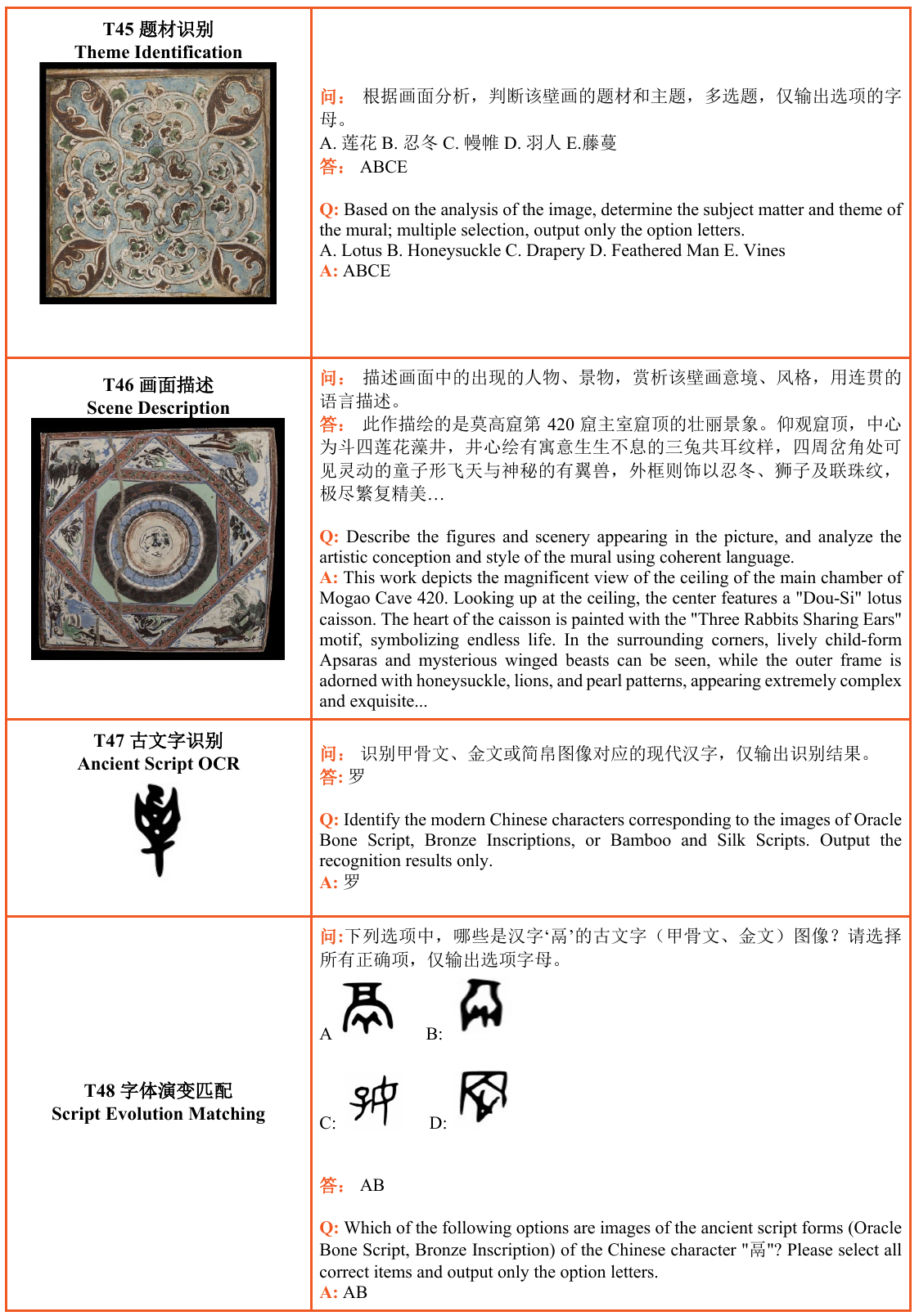}
    \caption{\benchmark{} examples for Tasks 45--48.}
    \label{fig:benchmark_dataset_example_13}
\end{figure*}

\begin{figure}[H]
    \centering
    \includegraphics[width=0.95\textwidth]{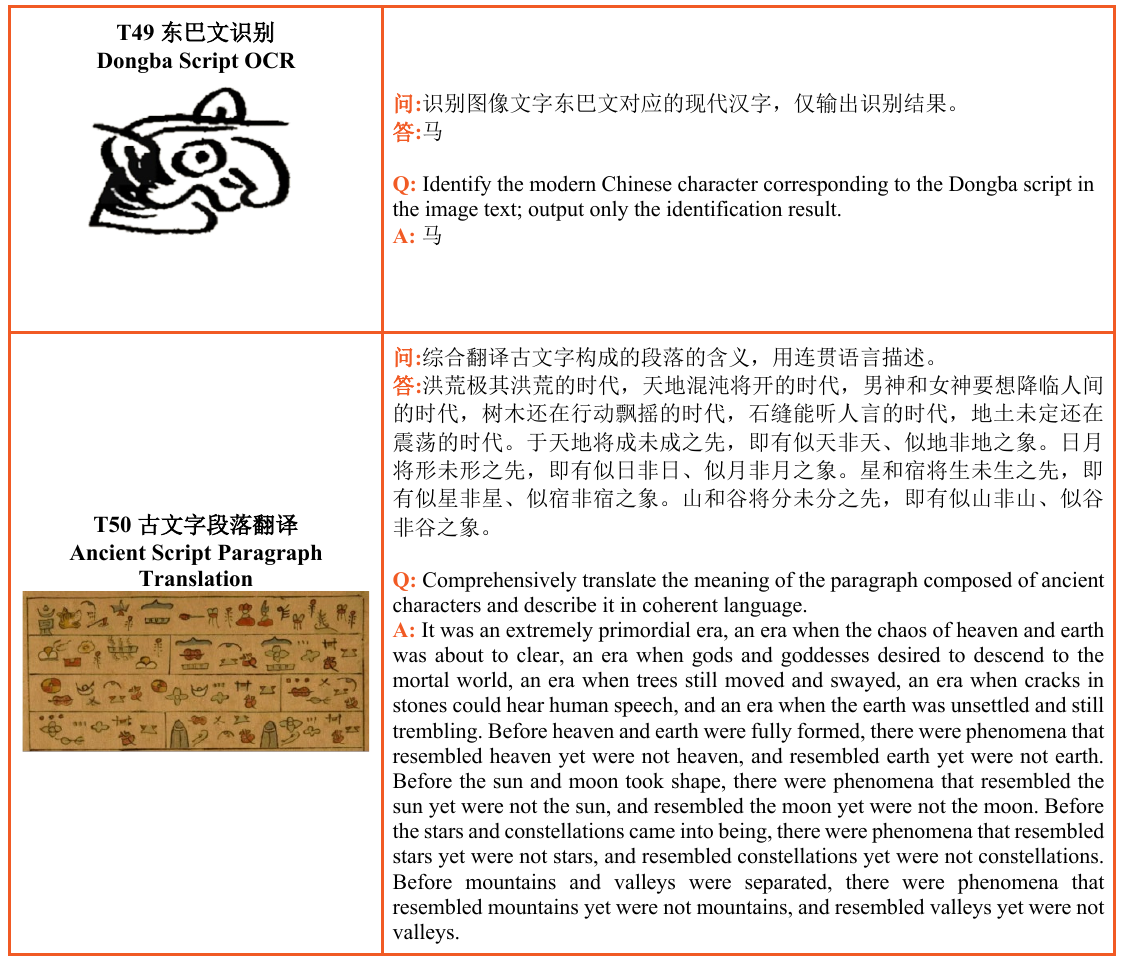}
    \caption{\benchmark{} examples for Tasks 49--50.}
    \label{fig:benchmark_dataset_example_14}
\end{figure}

Figure~\ref{fig:benchmark_subdomains} presents the seven CCH subdomains covered by \benchmark{}.
Detailed examples of the 50 tasks in \benchmark{} are presented in Figures~\ref{fig:benchmark_dataset_example_1}--\ref{fig:benchmark_dataset_example_14}.

\clearpage

\end{document}